\documentclass[10pt,twocolumn,letterpaper]{article}

\usepackage{iccv}
\usepackage{times}
\usepackage{epsfig}
\usepackage{graphicx}
\usepackage{amsmath}
\usepackage{amssymb}
\usepackage{booktabs}
\usepackage{caption}
\usepackage{subcaption}
\usepackage{enumitem}
\usepackage{xspace}

\usepackage[pagebackref=true,breaklinks=true,letterpaper=true,colorlinks,citecolor=blue,linkcolor=blue,bookmarks=false]{hyperref}

\iccvfinalcopy 

\def\iccvPaperID{} 

\newcommand{\ourtask}{2.5VRD\xspace}
\newcommand{\colorA}{\text{\textcolor{red}{A}}\xspace}
\newcommand{\colorB}{\text{\textcolor{blue}{B}}\xspace}

\newcommand\InsertImage[2][0.235]{
    \begin{subfigure}[c]{#1\linewidth}
        \centering
        \includegraphics[width=\textwidth]{#2}
    \end{subfigure}
}

\begin{document}

\title{2.5D Visual Relationship Detection}

\author{Yu-Chuan Su,
Soravit Changpinyo,
Xiangning Chen$^{1,2}$\thanks{Work done during an internship at Google.},
Sathish Thoppay,
Cho-Jui Hsieh$^2$,\\
Lior Shapira,
Radu Soricut,
Hartwig Adam,
Matthew Brown,
Ming-Hsuan Yang,
Boqing Gong\\
$^1$Google Research $\qquad$ $^2$UCLA
}

\maketitle
\ificcvfinal\thispagestyle{empty}\fi

\begin{abstract}

Visual 2.5D perception involves understanding the semantics and geometry of a scene through reasoning about object relationships with respect to the viewer in an environment.
However, existing works in visual recognition primarily focus on the semantics.
To bridge this gap, we study 2.5D visual relationship detection (2.5VRD), in which the goal is to jointly detect objects and predict their relative depth and occlusion relationships.
Unlike general VRD, 2.5VRD is egocentric, using the camera's viewpoint as a common reference for all 2.5D relationships.
Unlike depth estimation, 2.5VRD is object-centric and not only focuses on depth.
To enable progress on this task, we create a new dataset consisting of 220k human-annotated 2.5D relationships among 512K objects from 11K images.
We analyze this dataset and conduct extensive experiments including benchmarking multiple state-of-the-art VRD models on this task.
Our results show that existing models largely rely on semantic cues and simple heuristics to solve 2.5VRD, motivating further research on models for 2.5D perception.
The new dataset is available at~\url{https://github.com/google-research-datasets/2.5vrd}.

\end{abstract}

\section{Introduction} \label{sec:intro}

Visual 2.5D perception involves understanding the semantics and geometry of a scene: the relationships between objects with the viewer as the main reference point in an environment~\cite{humanvision}.
For instance, we may refer to a chair solely through semantics by its name and attributes (e.g., the wooden chair), through both semantics and geometry by its spatial relationship to other objects (e.g., the chair on the right of the table), or through the geometry only---by distance to our viewpoint (e.g., the chair that is closer). 
However, object recognition~\cite{deng2009imagenet}, detection~\cite{lin2014microsoft,kuznetsova2020open}, and segmentation~\cite{lin2014microsoft}, among other hallmark computer vision tasks, primarily focus on the semantics component.
As a result, most visual perception models operate in a 2D world and lack 2.5D visual understanding.

\begin{figure}
    \centering
    \includegraphics[width=\columnwidth]{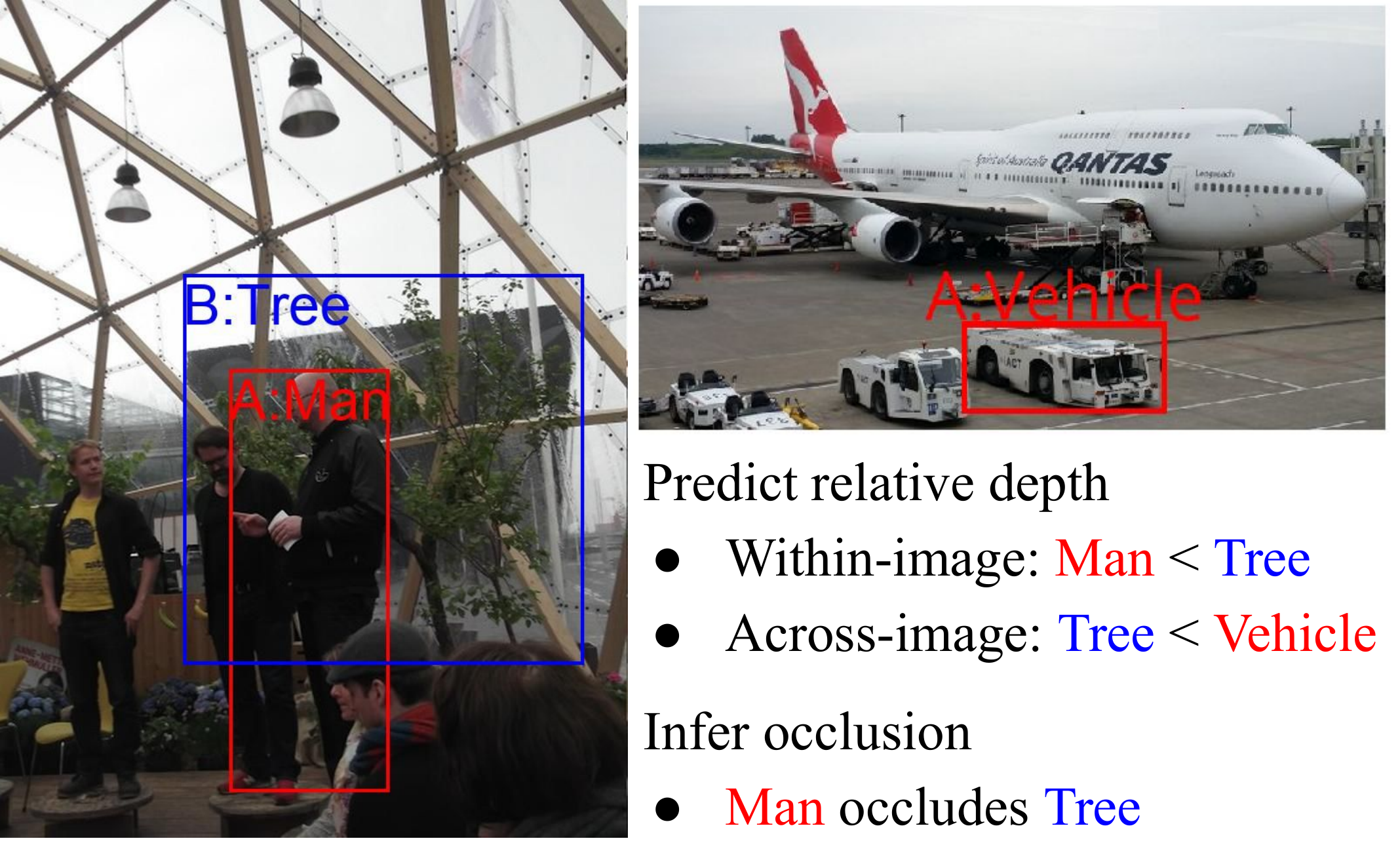}
    \vspace{-15pt}
    \caption{
    2.5D visual relationship detection (2.5VRD).
    We consider two relationships, i.e., relative depth and occlusion, both within an image and across images (best viewed in color; showing not all, but three objects).
    }
    \vspace{-9pt}
    \label{fig:teaser}
\end{figure}

Motivated by this, we introduce 2.5D visual relationship detection (\ourtask).
The goal of \ourtask is to detect objects and predict their relative depth and occlusion relationships as a unified task, as illustrated in Figure~\ref{fig:teaser}.
We study the relative depth in two settings: ``within an image'' and ``across images'' (e.g., the depth of the tree with respect to the man and to the vehicle.) Occlusion on the other hand only applies to the ``within an image'' setting. Clearly, to be able to perform well on this task, the geometry of a scene cannot be ignored.

Our task is primarily motivated by the scientific question ``Do machines possess 2.5D visual understanding capability like humans do?''
An answer to this question would benefit our understanding of machine visual perception.
Furthermore, we believe an effective model for the core problem of 2.5D visual relationships can benefit a wide range of applications, e.g., helping a self-driving vehicle to understand scenes beyond its LiDAR range, assisting a robot to navigate and manipulate objects, and improving (amodal) instance detection~\cite{gao2011segmentation,hsiao2014occlusion,zhang2018occlusion,kar2015amodal,deng2017amodal} and segmentation~\cite{li2016amodal,qi2019amodal,hu2019sail,russell2008labelme,jia2012learning,tighe2014scene,chen2015multi}, to name a few.

2.5VRD shares similar high-level motivation as visual relationship detection (VRD) and depth estimation, yet with important conceptual differences. Our task differs from \emph{general} VRD as it focuses on depth and occlusion. It also differs from recent work on \emph{spatial} VRD~\cite{sadeghi2011recognition,lu2016visual,yang2019spatialsense,krishna2017visual,kuznetsova2018open}. For example, the spatial relationships in SpatialSense~\cite{yang2019spatialsense} are concerned with both locations and poses of objects with respect to each other, while \ourtask is egocentric, defining occlusion and depth orders from the viewer's perspective.
Indeed, ``a chair (in SpatialSense) may be $behind$ a person even if it appears to the left of the person (depending on where that person faces)''.
The most similar work to ours is Rel3D~\cite{goyal2020rel3d} which also consists of view-dependent relationships, but, unlike ours, they are situated in synthetic environments.
Finally, the depth in \ourtask is object-centric, unlike the pixel-wise depth studied in monocular depth estimation~\cite{saxena2006learning,saxena2008make3d,liu2010single,silberman2012indoor,ladicky2014pulling,liu2015deep,eigen2015predicting,li2015depth,hane2015direction,shelhamer2015scene,xian2018monocular,lasinger2019towards,xian2020structure,chen2016single}.

To enable progress on our proposed \ourtask task, we introduce a new large-scale dataset of 219,570 2.5D relationships among 511,545 objects from 11,084 images of the Open~Images~\cite{kuznetsova2020open}. Our dataset is an order-of-magnitude larger than existing VRD datasets~\cite{lu2016visual,yang2019spatialsense}. It is also the first large-scale human-annotated dataset with 2.5D visual relationships (in two settings) on \emph{natural} images.
Additionally, unlike existing benchmarks, the annotations on our validation and test sets are exhaustive, allowing us to use both precision and recall as the evaluation metrics.

We analyze our dataset and conduct extensive experiments that shed light on the difficulty of \ourtask. First, we use the rich annotations to analyze how humans and visual recognition models tackle \ourtask.
We build a simple baseline and find that our baseline's performance and the agreement among five raters are both correlated well with the relative depth between two objects.
Second, we study the effect of various cues on the performance of our baseline model. Our results show that the object sizes and locations are important at predicting the relative depth, suggesting that high-quality object detection is key to 2.5VRD. On the other hand, the appearance cue is more important for occlusion prediction.
Finally, we benchmark four state-of-the-art VRD models on our 2.5VRD dataset. We find that they do not significantly outperform our simple baseline and that they do not generalize well from the within-image setting to the across-image setting.
These results suggest that existing models designed for 2D VRD are not sufficient for relative depth or occlusion reasoning.

In summary, our main contributions are as follows. We propose the 2.5VRD task, promoting the object-centric depth and occlusion reasoning as the first-class citizen. We concretize the task with an extensively labeled dataset, which is an order-of-magnitude larger than existing VRD datasets and unique in the exhaustive annotations on the validation and test sets. We propose a model to study various factors that may come into play, and we hope the findings will help design improved models in future. Finally, we evaluate four state-of-the-art VRD methods for 2.5VRD. Results show that their performance is comparable to the baseline model, highlighting the new challenges in 2.5VRD of which are not taken into account by these methods yet.

\section{Visual Relationships in 2.5D}

This section first formalizes the \ourtask task, followed by a detailed strategy for data and label collection. 
Next, we analyze the resulting dataset and study how humans approach \ourtask.
Finally, we compare \ourtask with related datasets and work.

\begin{figure}[t]
    \centering
    \includegraphics[width=1\columnwidth]{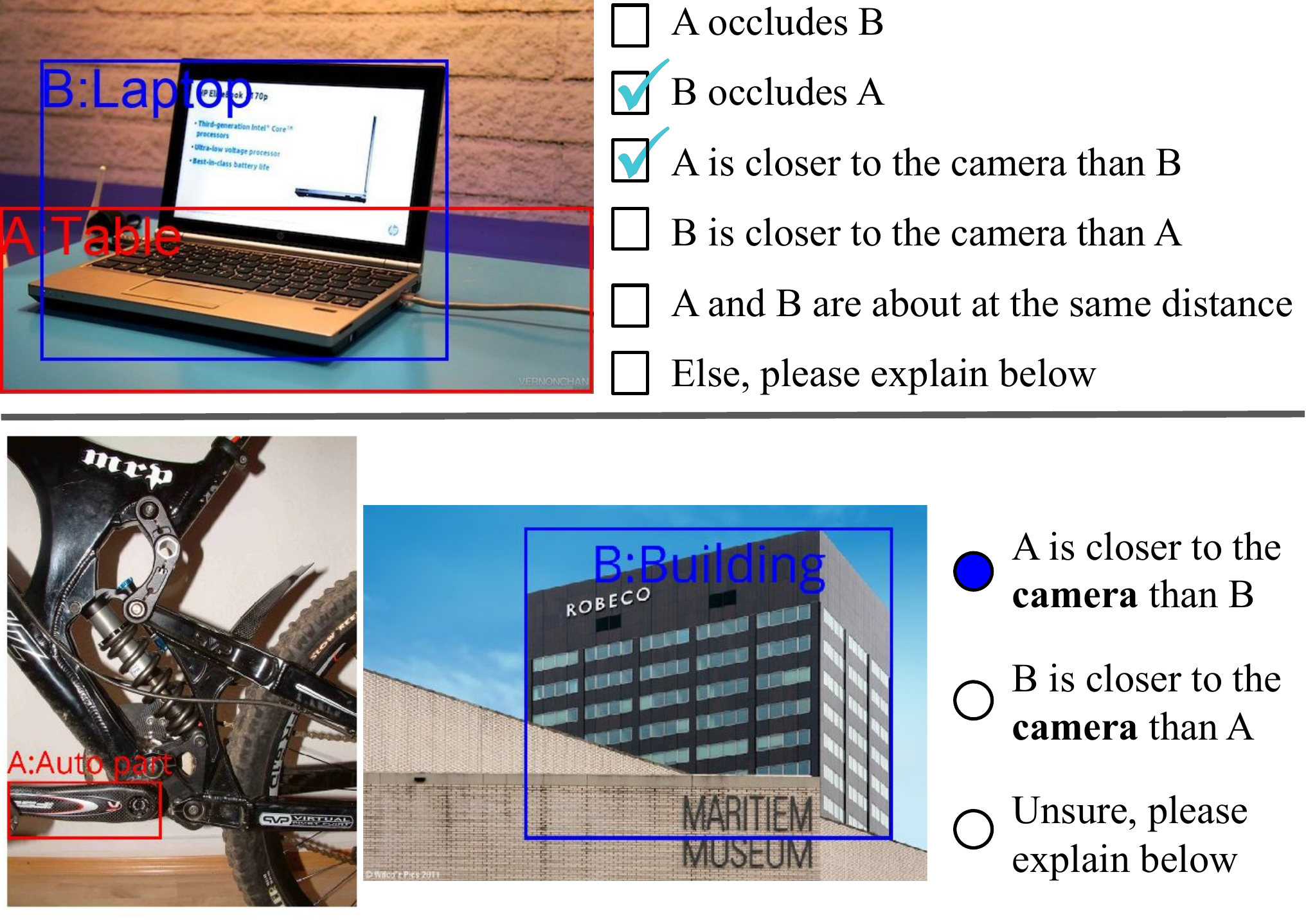}
    \vspace{-21pt}
    \caption{Annotation interface for within-image (top) and across-image (bottom) \ourtask.}
    \label{fig:UI}
    \vspace{-9pt}
\end{figure}

\subsection{Problem Formulation}
We formalize our task as follows.
Given two input images $(I_{a}, I_{b})$, the 2.5VRD task is to predict a set of 2.5D relationships.
Each 2.5D relationship consists of a triplet $\langle o_a, predicate, o_b \rangle$, where $o_{x}$ is an object in $I_{x}$ specified by a tight bounding box and its class name and $predicate$ is the relationship between $o_{a}$ and $o_{b}$.
By treating $I_{a}$ and $I_{b}$ as separate input, this formulation is applicable to both within-image and across-image setting.
For the within-image setting, i.e., $I_{a}$ and $I_{b}$ are identical, we consider both relative depth and occlusion relationships.
For the across-image setting, only the relative depth relationship is relevant.
Possible values for the $predicate$ include \{{\em is closer than, is farther than, is at the same depth as}\} for relative depth, where the $predicate$ ``{\em is at the same depth as}'' is on only in the within-image setting since we find it too challenging to label across images.
For occlusion, ${predicate}{\in}\{${\em occludes, does not occlude, mutual occlusion}$\}$.
Note that there could be no occlusion between two objects, and they could be mutually occluded.
Because the relationships are defined between any two objects,
there should be $N_{a}{\times}(N_{a}{-}1)$ and $N_{a}{\times}N_{b}$ relationships in within-image and across-image setting respectively, where $N_{x}$ is the number of objects in $I_{x}$.

One may alternatively formulate the problem for relative depth relationship as ranking of objects, but we find that some difficult pairs of objects often invalidate the ranking lists. It becomes especially troublesome when we merge the ranking lists from different raters. Hence, we instead use the $\langle o_a, predicate, o_b\rangle$ triplets considering their flexibility and the annotation cost.

\subsection{Data and Label Collection} \label{sec:data-label-collection}
We construct the \ourtask dataset on top of the Open Image Dataset (OID)~\cite{kuznetsova2018open}.
OID mostly consists of scenery images from Flickr, where each image may contain multiple objects and/or people.
We maintain the original train/validation/test split of the images. 
We use the annotated bounding boxes and 600 class names of objects in the OID images.
As the annotations in OID are over-complete (i.e., multiple bounding boxes for an object), we filter the boxes before collecting \ourtask labels.
We also ignore extremely small or large boxes (occupying less than $2\%$ or more than $70\%$ area of the image) to avoid ill-defined cases.
Finally, we remove boxes containing group of objects.
See supp.~for details.

\begin{table}[tp]
    \caption{Overview of the \ourtask dataset.} \label{tab:stats}
    \vspace{-9pt}
    \resizebox{\columnwidth}{!}{
        \begin{tabular}{clrrr}
            \multicolumn{2}{c}{} & Training & Validation & Test\\
            \cline{3-5}
            & Images  & 105,694 & 1,200 & 4,000\\
            & Objects & 493,498 & 4,063 & 13,893\\
            \midrule
            Within & Pairs of objects & 105,694 & 6,339 & 23,724\\
            \cline{2-5}
            image & $\colorA$ is closer &39.1\% & 39.9\% & 38.7\% \\
            &  $\colorB$ is closer  &39.9\% & 39.6\% & 38.9\% \\
            & Same depth & 10.1\% & 6.3\% & 7.5\% \\
            \cline{2-5}
            & \colorA occludes \colorB & 10.9\% & 10.3\%  & 9.3\% \\
            & \colorB occludes \colorA & 11.0\% & 10.0\%  & 9.1\% \\
            & Mutual occlusion & 3.5\% & 1.9\% & 2.1\%\\
            \midrule
            Across  & Pairs of images & 52,484 & 600 & 2,000\\
            image & Pairs of objects & 52,484 & 6,868 & 24,461\\
            \cline{2-5}
            & \colorA is closer  & 43.5\% & 40.9\% & 43.8\% \\
            & \colorB is closer & 45.2\% & 48.0\% & 44.5\% \\
            \bottomrule
        \end{tabular}
    }
    \vspace{-6pt}
\end{table}

\paragraph{Labeling \ourtask within an image.} For each training image, we have a rater to label one randomly formed pair of objects. 
The label consists of both relative depth and occlusion relationships (see the top panel in Figure~\ref{fig:UI}, which is a screenshot of the annotation UI).
We also include an ``unsure'' option for ambiguous cases in light of the difficulty of the problem.
For each validation or test image, we collect five \ourtask labels for every pair of objects from five raters respectively. 
We then use majority voting to determine the final labels. 
This strategy ensures that the labels for the validation and test set are of high quality.
It also results in comprehensive annotations for all pairs of objects in a validation or test image, allowing us to evaluate model performance in terms of precision and recall. 
Note that we annotate only one pair of objects in the training set to maximize the number of training samples under the budget constraint, based on the hypothesis that higher diversity in the training data is important for model performance.

\paragraph{Labeling \ourtask across images.} We split the training images into two groups and then construct pairs by selecting one image from either group. 
Given a pair of images, we randomly choose an object from each of them. 
A rater ranks the two objects by their depths using the annotation UI illustrated by the bottom panel in Figure~\ref{fig:UI}.
We pair up validation and test images in the same way, but provide dense labels for all across-image object pairs. 
In addition, we assign each of them to five raters to secure high-quality labels for the validation and test sets.

\begin{table}[t]\small
    \centering
    \tabcolsep=0.1cm
    \caption{Distributions of difficulty scales for within-image and across-image object depth ordering, respectively.}
    \label{tab:difficulty-scales}
    \vspace{-4pt}
    \begin{tabular}{cccccc}
                      & Easy & Moderate & Difficult & Infeasible & Ambiguous\\
        \midrule
        Within-image  & 55.8\% & 16.4\% & 13.1\% & 10.5\% & 4.3\% \\
        Across-image & 50.0\% & 21.6\% & 16.9\% & 6.7\% & 4.8\%\\
        \bottomrule
    \end{tabular}
    \vspace{-6pt}
\end{table}

\subsection{Dataset Statistics and Analyses} \label{sec:stats}
Table~\ref{tab:stats} shows the statistics for the proposed \ourtask dataset.
Note that the within-image and across-image setting share the common sets of images and objects within each split.
Out of the within-image object pairs, about 80\% have apparent depth disparities (rows of ``$\colorA$ is closer'' and ``$\colorB$ is closer''), and approximately 10\% are at about the same depth.
Furthermore, one object occludes the other (but not vice versa) in about 20\% of the pairs (rows of ``\colorA occludes \colorB'' and ``\colorB occludes \colorA''), and about 3\% are mutually occluded. 
For across-image \ourtask, the raters managed to tell the difference between two objects' depths for approximately 88\% of all the pairs (see the last two rows in the table). 
The dataset preserves the object pairs for which the raters selected the unsure option, as learning models may make sense of them in the future.

\begin{figure*}[t]
    \centering
    \includegraphics[width=\linewidth]{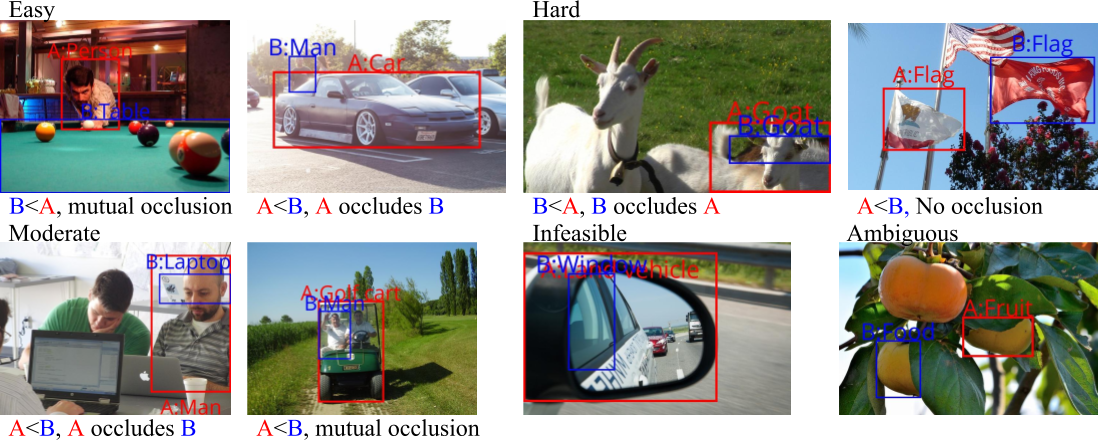}
    \vspace{-15pt}
    \caption{Examples of within-image \ourtask with different difficulties. \textcolor{red}{A}$<$\textcolor{blue}{B} means that \textcolor{red}{A} is closer to the viewpoint than \textcolor{blue}{B}.}
    \label{fig:examples-in-image}
\end{figure*}

\begin{figure}
    \centering
    \includegraphics[width=\columnwidth]{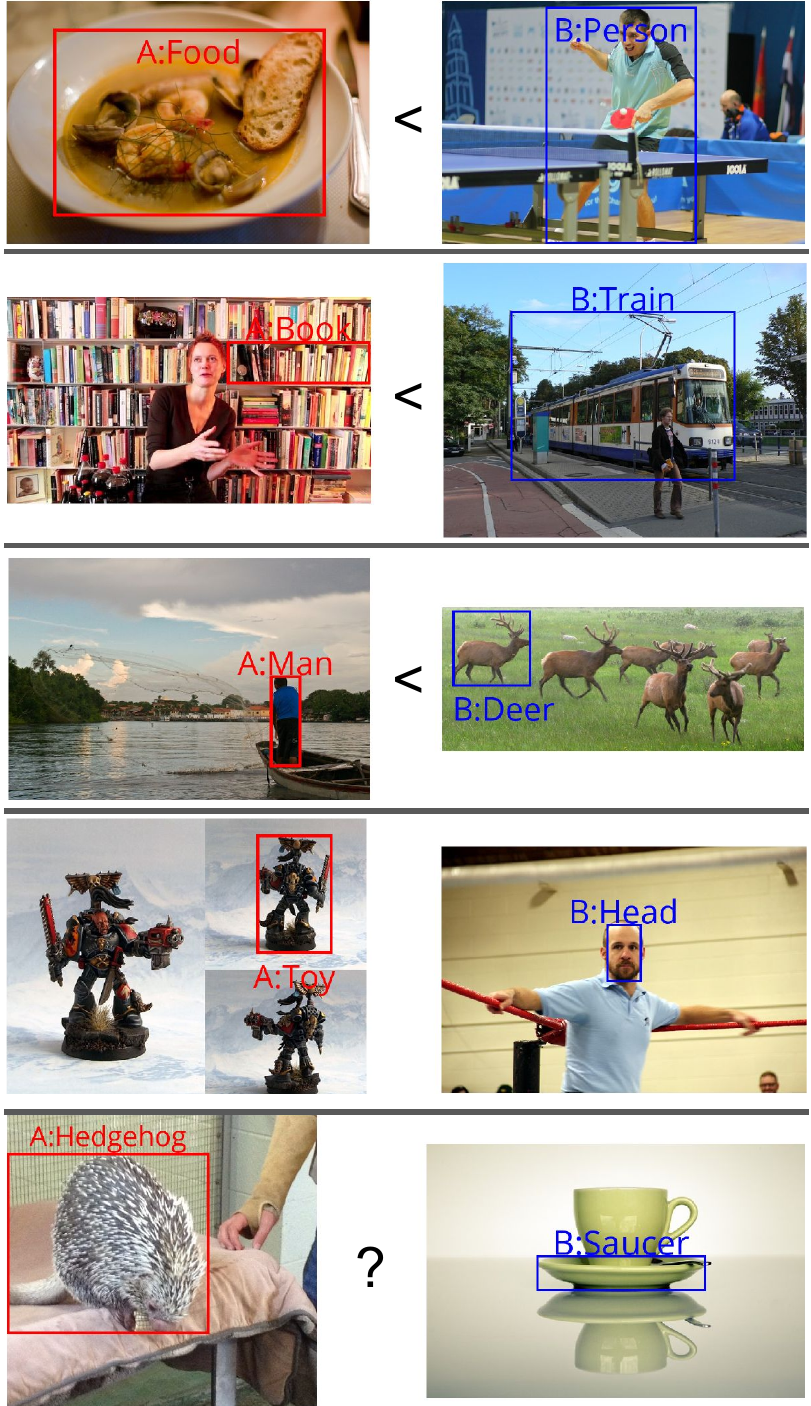}
    \vspace{-15pt}
    \caption{Examples of across-image \ourtask. 
    The difficulty is easy, moderate, difficult, infeasible, and ambiguous from top to bottom.
    }
    \label{fig:examples-ximg}
    \vspace{-3pt}
\end{figure}

\paragraph{Human perception of \ourtask.} The annotations on each validation or test example by five raters allow us to analyze how humans approach the \ourtask task. 
We define five difficulty scales for depth ordering:
\begin{description}  \setlength\itemsep{0pt}
\item[Easy:] Five raters all agreed on a relative depth label and did not choose the unsure option.
\item[Moderate:] Four out of five raters agreed with each other.
\item[Difficult:] Three out of five agreed on a relative depth label. 
\item[Infeasible:] A majority of the raters chose ``unsure''. 
\item[Ambiguous:] There is no majority agreement on any label. 
\end{description}
Table~\ref{tab:difficulty-scales} shows the distribution of validation and test examples over the five difficulty scales. 
For the proposed dataset, more than 50\% of the object pairs belong to the ``easy'' scale. 
Moreover, more within-image object pairs fall into the ``easy'' scale than the across-image object pairs, likely because the latter requires the raters to estimate metric depths to some extent. 
In contrast, relative depths are sufficient to rank two objects in the same scenery image. 
There are 7\% and 10\% infeasible object pairs for across-image and within-image \ourtask, respectively, meaning that a majority of the raters were ``unsure'' how to rank them by depth. 
Finally, less than 5\% of the object pairs received no label because there was no majority winner. 
Overall, the \ourtask task is more difficult for humans than we expected, considering that a notable proportion of examples are ``infeasible'' or ``ambiguous'' for human raters to reach a consensus.

Figure~\ref{fig:examples-in-image} and Figure~\ref{fig:examples-ximg} show some examples and their labels (more in the supplementary materials). 
There is a high (negative) correlation between the difficulty scales and the object pairs' depth differences. 
The raters chose ``unsure'' or became ambiguous about some object pairs mainly for the following reasons. 
Two objects could appear at about the same depth. 
One or both lack backgrounds for the raters to infer depths. 
The images may not be natural scenes (e.g., edited images, paintings, cartoons, etc.). 

\begin{figure*}
    \centering
    \includegraphics[width=0.45\textwidth]{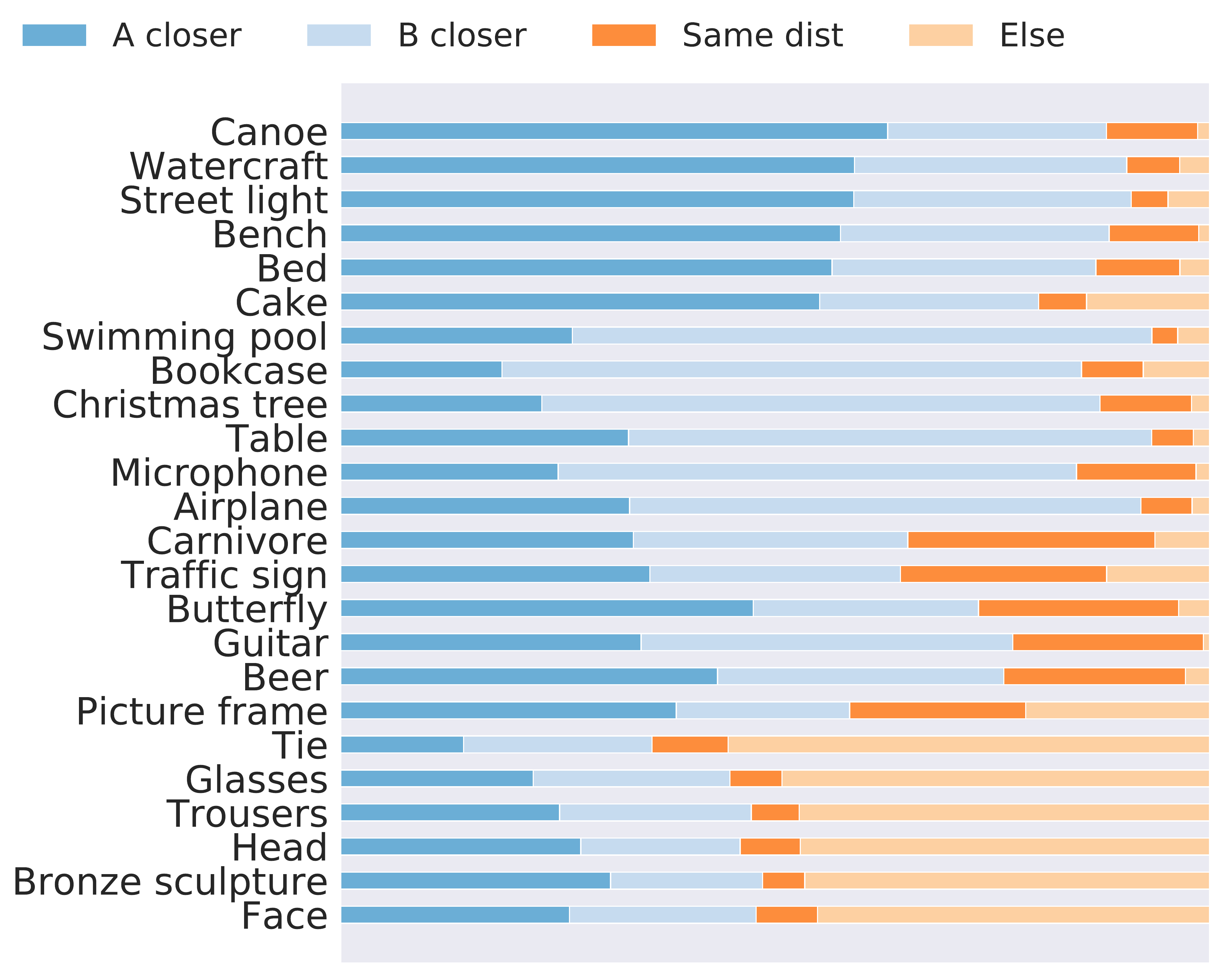}
    \includegraphics[width=0.53\textwidth]{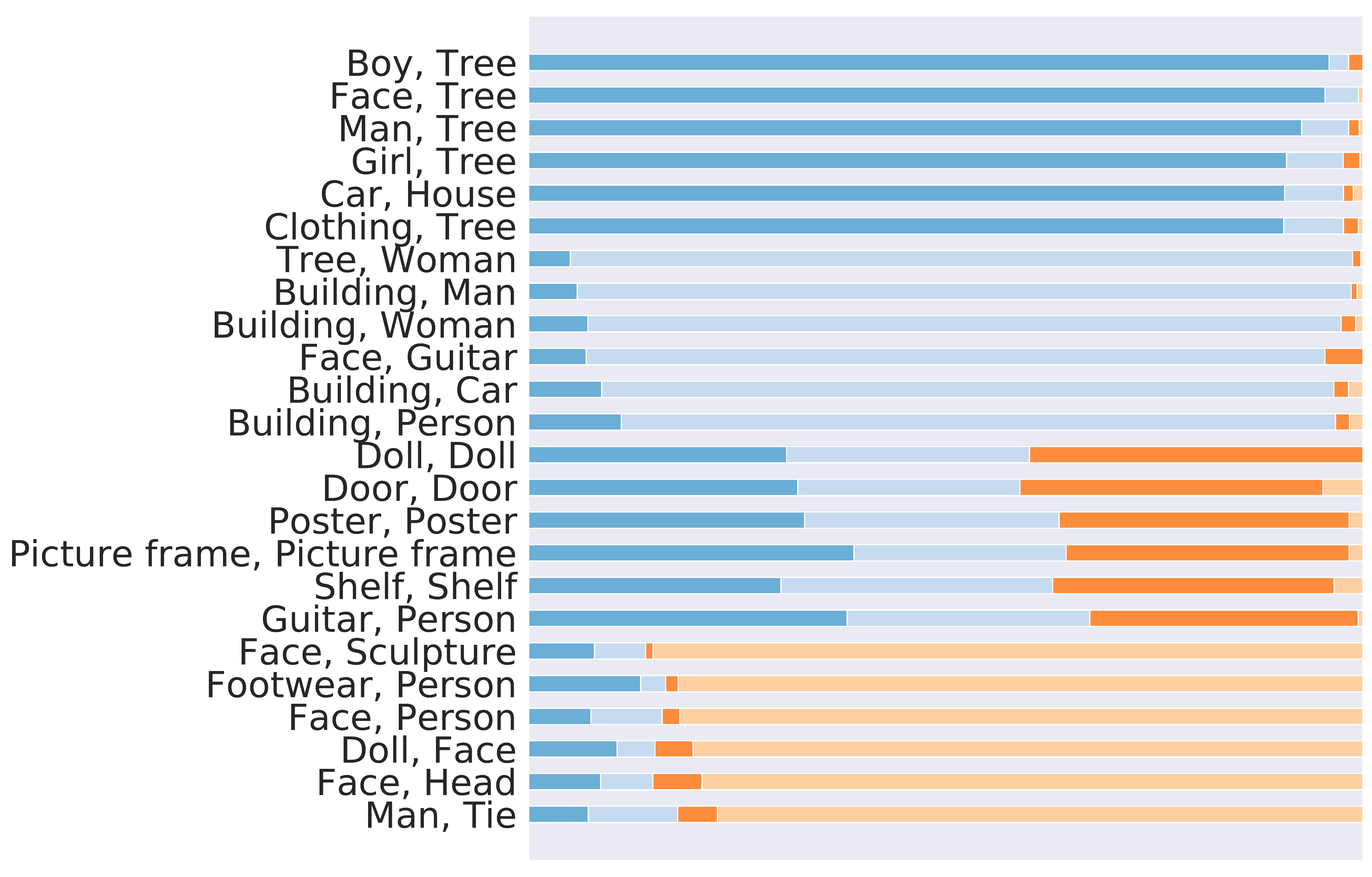}
    \vspace{-12pt}
    \caption{Distributions of depth labels given object classes (Y-axis: Object A) or object pairs (Y-axis: (Object A, Object B)).}
    \label{fig:depthdist}
    \vspace{-9pt}
\end{figure*}

\vspace{-10pt}
\paragraph{Potential bias.} We visually inspect each object class and its relative depth label distribution, as well as each object pair and its relative depth label distribution. 
For simplicity, we focus on the within-image setting (and ignore the across-image examples). 
Figure~\ref{fig:depthdist} shows the top six object classes (and object pairs) with the highest percentage for each label. 
Take the left panel in the figure for example. 
The first six rows correspond to the most frequent six classes that are ``closer than B'', the next six rows are the most frequent classes which ``B are closer than'', and so on. 
We observe a natural bias. 
For example, big and background objects such as swimming pool, bookcase, and tree tend to be further than the other objects. 
Further, many object pairs of the same class (e.g., dolls, doors, and posters) are of the same distance, with man and guitar being two exceptions. 
Finally, clothes, person-like objects, and body parts are ambiguous categories. 
We do not attempt to correct the natural bias as it is a reflection of our daily scenes.
The supplementary materials contain a similar study about the occlusion labels.

\begin{table*}
\caption{\ourtask vs.\ existing datasets ({*}2.5D relations
for existing datasets have
``behind'' or ``front'' in their $predicates$)} \label{tab:data-comparison}
\vspace{-6pt}
\resizebox{\textwidth}{!}{
\begin{tabular}{lrrrrrrrr}
 & Images & Objects & Classes & Predicates & Relations & 2.5D relations{*} & Occlusion & Across-img relations\tabularnewline
\midrule
VRD~\cite{lu2016visual} & 5,000 & 32,901 & 100 & 70 & 37,993 & 4,780 & 0 & 0\tabularnewline
SpatialSense~\cite{yang2019spatialsense} & 11,569 & 33,861 & 3,679 & 9 & 17,498 & 5,132 & 0 & 0\tabularnewline
VG (relationship)~\cite{krishna2017visual} & 108,077 & 2,254,357 & 65,405 & 4,016 & 2,316,104 & 66,660 & 0 & 0\tabularnewline
OID (relationship)~\cite{kuznetsova2018open} & 596,308 & 1,478,971 & 303 & 31 & 3,284,282  & 0 & 0 & 0\tabularnewline
\ourtask (ours) & 110,894 & 511,454 & 600 & 5 & 219,570 & 219,570 & 29,105 & 83,813\tabularnewline
\bottomrule
\end{tabular}
}
\end{table*}

\subsection{Related Datasets and Work}

\paragraph{VRD.} Sadeghi and Farhadi studied VRD using 17 unique relationships~\cite{sadeghi2011recognition}. Lu et al.~scaled up the study by a new benchmark with 37,993 relations over 5,000 images~\cite{lu2016visual}. 
They showed that language prior was effective for detecting the visual relations with few to no training examples. 
Peyre et al.~collected 76 unusual relationships to evaluate model generalization for VRD~\cite{peyre2017weakly}. 
SpatialSense curated 17,498 spatial relationships over 11,569 images~\cite{yang2019spatialsense}. 
The Visual Genome (VG)~\cite{krishna2017visual} and OID~\cite{kuznetsova2018open} provide VRD labels albeit sparse per image. 

Table~\ref{tab:data-comparison} contrasts our \ourtask dataset with the related, representative datasets. VRD and SpatialSense are arguably the most widely-used benchmarks for VRD. 
Our dataset is an order of magnitude larger than them in the numbers of images, objects, and relations. 
While \ourtask makes depth and occlusion the first-class citizen, almost all relationships in the existing VRD datasets are 2D. 
Only 5,132 relations in SpatialSense have ``behind'' or ``front'' in their $predicates$, and yet a cell phone could be ``in front of'' a person as long as the person faces to the phone even if the phone is farther to the viewpoint. 
The same issue occurs in the VG dataset, where the  ``behind'' relationship mostly refers to an object's orientation, rather than depth.
Across-image \ourtask is unique in our dataset, made possible due to the relative depth $predicates$ between objects.

Besides the existing VRD datasets, our work is also closely related to the rich line of VRD methods and models~\cite{liang2017deep,zhang2017ppr,zhang2017relationship,zhuang2017towards,li2017vip,plummer2017phrase,zhang2017visual,ma2018attend,dai2017detecting,yao2018exploring,yang2018graph,zhang2019large,woo2018linknet,yu2017visual,peyre2017weakly,yin2018zoom}, which will speed up tackling \ourtask. We leave the exploration into them to future work. Instead, our experiments aim to help readers gain more insights into \ourtask, especially about how different visual cues interplay in the 2.5D visual relationships. 
Our work is also related to human-object interactions~\cite{gupta2009observing,yao2010grouplet,delaitre2011learning,kato2018compositional,gkioxari2018detecting,chao2015hico,qi2018learning,chao2018learning,zhuang2017care}, which may be viewed as human-centric VRD. In contrast, \ourtask is  egocentric, using the viewpoint as the reference for the relationships between two arbitrary objects. 

\vspace*{-0.1in}
\paragraph{2.5D perception.} 
Monocular depth estimation~\cite{saxena2006learning,saxena2008make3d,liu2010single,silberman2012indoor,ladicky2014pulling,liu2015deep,eigen2015predicting,li2015depth,hane2015direction,shelhamer2015scene,chen2016single,xian2018monocular,lasinger2019towards,xian2020structure} infers a dense, pixel-wise depth map from an image, and thus not object-centric.
Our empirical results show that, while dense depth maps provide informative cues to \ourtask,
it is far from solving depth ordering for objects and does not account for occlusion.
Unlike depth, existing works mostly do not consider occlusion as an independent task, but a latent factor to improve object detection~\cite{gao2011segmentation,hsiao2014occlusion,zhang2018occlusion}, semantic and instance segmentation~\cite{russell2008labelme,jia2012learning,tighe2014scene,chen2015multi}, and other applications~\cite{jiang2020peek,zhan2020self}. Instead, \ourtask directly deals with occlusion.

Some works study occlusion and depth ordering between image regions instead of objects~\cite{sundberg2011occlusion,hoiem2011recovering,zhang2015monocular,jia2012learning,lu2019occlusion,zhu2017semantic,qi2019amodal}.
Because they define occlusion along the object or scene boundaries, the relationship is always binary.
Also, most of them couple the two relationships and define depth order based on occlusion~\cite{jia2012learning,lu2019occlusion,zhu2017semantic,qi2019amodal}, so the relative depth is only defined within a connected component where the regions overlap.
While some works define depth order globally, they rely on the existence of the ground~\cite{hoiem2011recovering} or 3D object bounding boxes~\cite{zhang2015monocular} and consider only objects on the ground or cars.
In contrast, we consider the 2.5D relationships between arbitrary objects, leading to a more general relationships definition and a larger dataset.

Our work is also broadly related to amodal instance segmentation~\cite{li2016amodal,zhu2017semantic,qi2019amodal,hu2019sail} and amodal object detection~\cite{kar2015amodal,deng2017amodal}.
We envision that detecting depth and occlusion relationships between objects can facilitate amodal tasks and vice versa.
Finally, single-view 3D object detection~\cite{weiss2001model,geiger2012we,xiang2014beyond,song2015sun,chen2016monocular,mousavian20173d,ma2019accurate,liu2019deep,li2019gs3d,huang2019perspectivenet,brazil2020kinematic,Chen_2020_CVPR} is related but requires labor-intensive data collection, limiting existing work to mainly indoor and self-driving environments. 

\section{Experiments and Analyses}

In this section, we evaluate the performance of visual recognition models on 2.5VRD.
The goal of our experiments is to understand the effect of different visual signals and models on the 2.5VRD performance and establish the baseline results for future work.
To this end, we develop two baselines and benchmark them along with four state-of-the-art VRD methods on our proposed 2.5VRD task. Code and data will be made publicly available.

\subsection{Approaches to \ourtask}

In our experiments, we explore a two-stage approach for this task.
The first stage leverages an oracle/off-the-shelf object detectors to provide/infer multiple $(o_{a}, o_{b})$ pairs.
Then, given $(I_{a}, o_{a}, I_{b}, o_{b})$ as input, we infer 2.5D relationships between $o_a$ and $o_b$ by predicting the $predicate$ for each relationship.
Our baseline and state-of-the-art models will operate in this second stage.
Because 2.5D relationships are directional, we treat $\langle o_a$, {\em occludes}, $o_b \rangle$ and $\langle o_b$, {\em occludes}, $o_a \rangle$ as two different labels for $(o_{a}, o_{b})$.
We also include the ``unsure'' label for the relative depth relationship.
This leads to four possible values for each 2.5D relationship.

\paragraph{Overview of visual cues}
All baselines and state-of-the-art models explored in this paper employ a subset of the following four types of visual cues. The first one is direct semantics in the form of object class labels. For example, a person is often closer to the viewpoint than trees and buildings; more examples are provided in Figure~\ref{fig:depthdist}. The second cue is the geometric cue in the form of box size and location. For example, an overlap implies a probable occlusion relation. The third cue is appearance, both in term of object and its context. The forth cue is depth, both in terms of object and its context.

\subsubsection{Rule-Based Baselines}

We explore the following rule-based baselines, each of which relies on a specific visual cue.
\begin{itemize}[leftmargin=*,label=$\bullet$]
    \setlength{\itemsep}{2pt}
    \setlength{\parskip}{2pt}
    \item \textbf{Object class} predicts the most frequent $predicate$ for the pair of object classes in the training set.
    \item \textbf{Size} predicts $o_a$ is closer to the camera than $o_b$ if $o_a$'s box size is larger by a margin $\Delta_s$ (based on the fact that an object's size in an image is inversely proportional to its depth.)
    For occlusion prediction, if the size of the overlap area is larger than a threshold, the object that is closer occludes the other; otherwise, no occlusion.
    \item \textbf{Location}
    predicts $o_a$ is closer to the camera than $o_b$ if $o_a$'s Y-coordinate is larger by by a margin $\Delta_l$.
    For occlusion prediction, we couple the rule with relative depth prediction as in \textbf{Size}.
    \item \textbf{Depth}
    For depth prediction, we assume a depth map produced by a monocular depth estimator MiDaS~\cite{lasinger2019towards} and compute a depth estimate $D_a$ for each object by averaging the depth values inside its bounding box\footnote{We explored different methods for combining the inferred depth values but did not observe significant differences.}.
    $o_a$ is closer to the camera than $o_b$ if $D_a$ is smaller than $D_b$ by a margin $\Delta_d$
    For occlusion prediction, we again couple the rule with relative depth prediction as in \textbf{Size} and \textbf{Location}.
\end{itemize}
The margins are set to $\Delta_{s}{=}0.0$, $\Delta_l{=}0.02$, and $\Delta_{d}{=}0.02$, respectively, by a grid search on the validation set.

\subsubsection{Simple MLP Baselines}

We explore a two-layer multi-layer perceptron (MLP) followed by two heads, treating depth and occlusion predictions as two multi-class classification problems. This model takes in up to four types of visual signals, as detailed below.

\begin{itemize}[leftmargin=*,label=$\bullet$]
    \setlength{\itemsep}{2pt}
    \setlength{\parskip}{2pt}
    \item \textbf{Object class feature} represents an object's class using a one-hot vector. 
    \item \textbf{Bounding box feature} uses the bounding boxes' coordinates, concatenated with the overlap region's height, width, and area.
    \item \textbf{Appearance feature} extracts the appearance feature for an object from the bounding box locally and from the image globally using a Faster-RCNN~\cite{ren2015faster} pre-trained on OID~\cite{kuznetsova2018open} with an Inception-ResNet~\cite{szegedy2016inception} backbone.
    Given an image, we first obtain a feature map from Faster-RCNN's last convolutional layer. We then perform average pooling over the feature map to obtain the image appearance feature and ROI pooling over a bounding box to reap the corresponding object's appearance feature. Concatenating the image feature and the object feature provides information about the object's surroundings and the object itself. 
    \item \textbf{Depth feature} extracts depth information using MiDaS~\cite{lasinger2019towards}.
    Given the depth map (without per-image normalization), we compute the mean, standard deviation, minimum, and maximum of the depth values within the bounding box of an object as the depth feature. Further, we also compute the depth feature for the entire image and concatenate it with the object's depth feature.
\end{itemize}

\vspace*{-0.1in}
\paragraph{Implementation details}
We use concatenation to combine features from a pair of objects and to combine features of different types. We use a hidden layer of size 1024. We use the sum of cross-entropy losses over the two classification heads. We augment each training input with $(I_a, o_a, I_b, o_b)$ with $(I_b, o_b, I_a, o_a)$, and also randomly perturb the center, width, and height of bounding boxes by 10\% in addition to other standard perturbation to the images' saturation, contrast, brightness, and hue during training. 
The model is trained using Adam for 60,000 steps with a base learning rate of $2{\times}10^{-4}$ and a batch size of 32. We add an $L_{2}$ regularization with weight $1{\times}10^{-4}$ and use dropout with ratio $0.5$.

\subsubsection{State-of-the-art VRD Methods}
We explore the following state-of-the-art VRD models.
\begin{itemize}[leftmargin=*,label=$\bullet$,topsep=2pt]
    \setlength{\itemsep}{2pt}
    \setlength{\parskip}{2pt}
    \item \textbf{ViP-CNN}~\cite{li2017vip} predicts predicates using visual features from three bounding boxes, including the two object bounding boxes and a tight bounding boxes covering the union of the two objects.
    \item \textbf{PPR-FCN}~\cite{zhuang2017towards} is similar to \textbf{ViP-CNN} but adopts a different architecture to combine the information.
    \item \textbf{DRNet}~\cite{dai2017detecting} takes as input the appearance feature, location, and word vector embedding of two objects. The model architecture is designed by unrolling a conditional random field model.
    \item \textbf{VTransE}~\cite{zhang2017visual} predicts predicates from the feature vector difference between two objets,
    where the features involve appearance, location, and word vector embedding.
\end{itemize}
See supp.~for details.
Note that \textbf{ViP-CNN} and \textbf{PPR-FCN} take the union of two boxes as input and are therefore not applicable for the cross-image task.

\begin{table}[t]\small
    \centering
    \tabcolsep=0.12cm
    \caption{
        2.5VRD results of rule-based (top), MLP (middle), state-of-the-art visual relationship detection (bottom) models. Both rule-based and MLP use different visual cues.
        For MLP, B: bounding box feature, C: object class feature, D: depth feature, A: appearance feature. Best numbers in {\bf bold} and second-best in \underline{\textit{underlined and italic}}.
    }
    \label{tab:accuracy}
    \vspace{-5pt}
    \resizebox{\columnwidth}{!}{%
    \begin{tabular}{lccc|c}
                 & Within-image & Occlusion & Across-image & Average \\
    \midrule
    Rule: Object class     & 0.011 & 0.133 & 0.025 & 0.056\\
    \underline{\it Rule: Location} & \underline{\it 0.286} & {\bf 0.303} & 0.214 & \underline{\it 0.268}\\
    Rule: Size     & 0.232 & {\bf 0.303} & \underline{\it 0.240} & 0.258\\
    {\bf Rule: Depth}        & {\bf 0.292} & {\bf 0.303} & {\bf 0.303} & \bf{0.299}\\
    \midrule
    MLP: B            & 0.232 & 0.308 & 0.243 & 0.261\\
    MLP: B+C          & 0.280 & 0.317 & 0.314 & 0.304\\
    MLP: B+D          & 0.301 & 0.308 & 0.326 & 0.312\\
    \underline{\it MLP: B+A}          & \underline{\it 0.307} & \underline{\it 0.320} & \underline{\it 0.367} & \underline{\it 0.331}\\
    {\bf MLP: B+C+D+A}      & {\bf 0.310} & {\bf 0.324} & {\bf 0.370} & {\bf 0.335}\\
    \midrule
    ViP-CNN~\cite{li2017vip} & 0.336 & 0.342 & - & - \\
    PPR-FCN~\cite{zhuang2017towards} & 0.335 & 0.339 & - & - \\
    DRNet~\cite{dai2017detecting} & 0.338 & 0.344 & 0.366 & 0.349 \\
    VTransE~\cite{zhang2017visual} & 0.324 & 0.329 & 0.365 & 0.339 \\
    \bottomrule
    \end{tabular}
    }
\end{table}

\subsection{Evaluation Metrics}
We can evaluate a model's performance by precision and recall since we exhaustively label all pairs of objects in an image or between images of the validation and test sets. We report F1-scores in the main text and all metrics in the supplementary materials. 

A \ourtask model detects objects and predicts $predicate$ between any pair of them. We first use the same filtering procedure in data collection (see Section~\ref{sec:data-label-collection}) to discard extremely small or big box and ill-defined relations. Supposing that $N$ objects in an image survive this procedure, there will be $N{\times}(N-1)$ 2.5D visual relationships---for each pair of objects,
we predict two $predicates$ for both directions, respectively, because the depth and occlusion relationships are directional. To compute precision and recall, we find true positive 2.5D visual relations as follows. A detected relationship, $\langle o_{a}, predicate, o_{b}\rangle$, is considered correct if it satisfies two conditions. 
1) Both objects $o_{a}$ and $o_{b}$ are detected correctly. We consider $o_a$ as correct detection if it has greater than 0.5 intersection-over-union with the groundtruth box. 2) The predicted $predicate$ is correct. Similarly, we use the F1-score to evaluate across-image \ourtask, where the object pairs are between images.

We shift the evaluation metrics toward the quality of the $predicate$, not the object detection, by making no requirement over the predicted class for a detected bounding box. We further ``leak'' some information to the detector so that it keeps $N$, the number of bounding boxes after the filtering procedure, the same as the number of groundtruth objects in an image. Compared with the commonly used average precision metric in object detection, F1-score allows us to evaluate a model's \ourtask performance using either groundtruth or detected objects, facilitating us to analyze the sources of error in the \ourtask models.

\subsection{Results and Analyses}
We use the Faster-RCNN detector pre-trained on OID to detect objects for all the experiments except in Table~\ref{tab:tasks_accuracy}, where we employ the groundtruth bounding boxes.
We present more experiments and analysis in the supp.

\begin{figure}[t]
    \vspace{-4pt}
    \centering
    \begin{subfigure}[t]{0.48\linewidth}
        \centering
        \includegraphics[width=\linewidth]{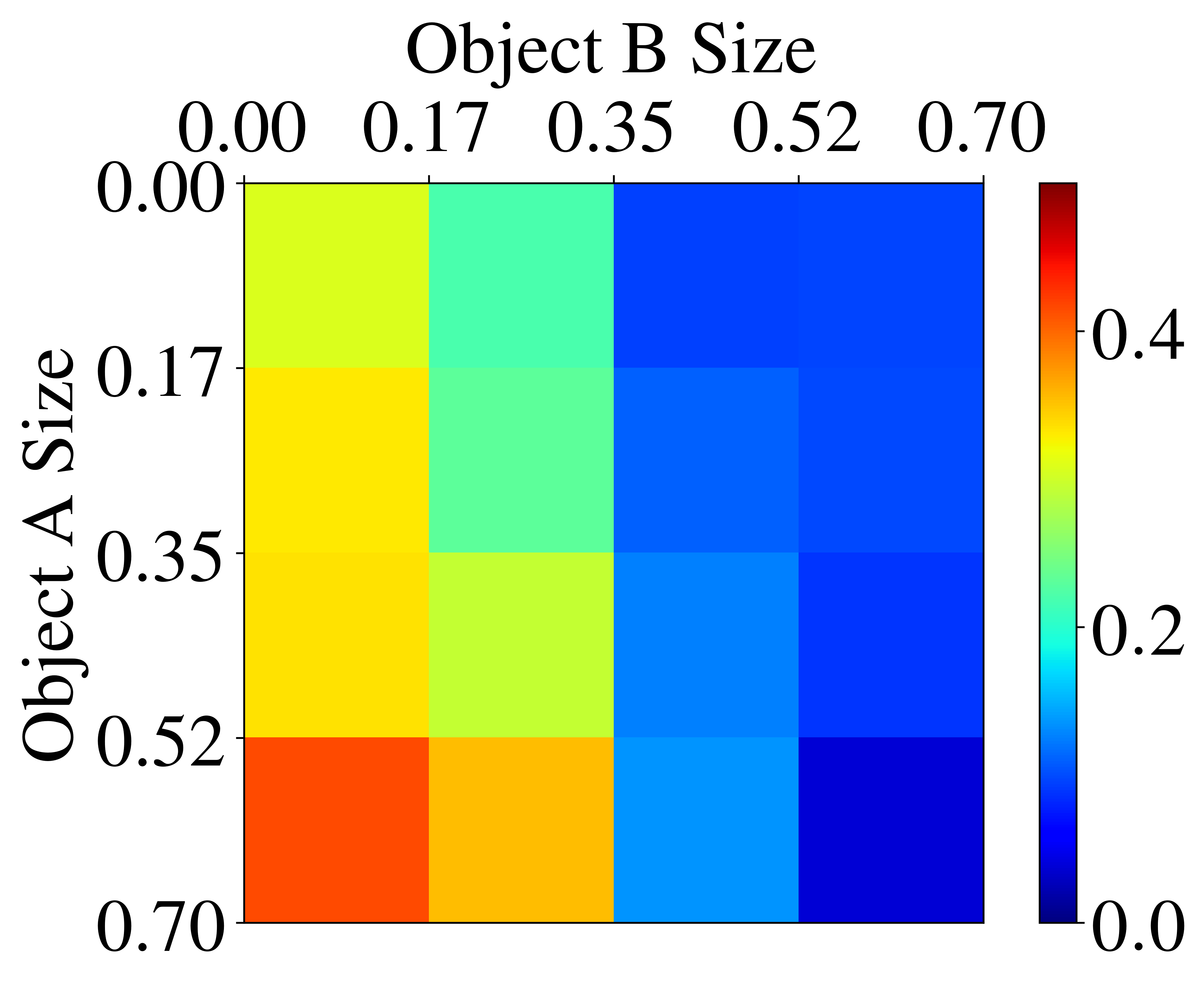}
        \vspace{-21pt}
        \caption{Within-image}
        \label{fig:withinimage_objectsize}
    \end{subfigure}
    \begin{subfigure}[t]{0.48\linewidth}
        \centering
        \includegraphics[width=\linewidth]{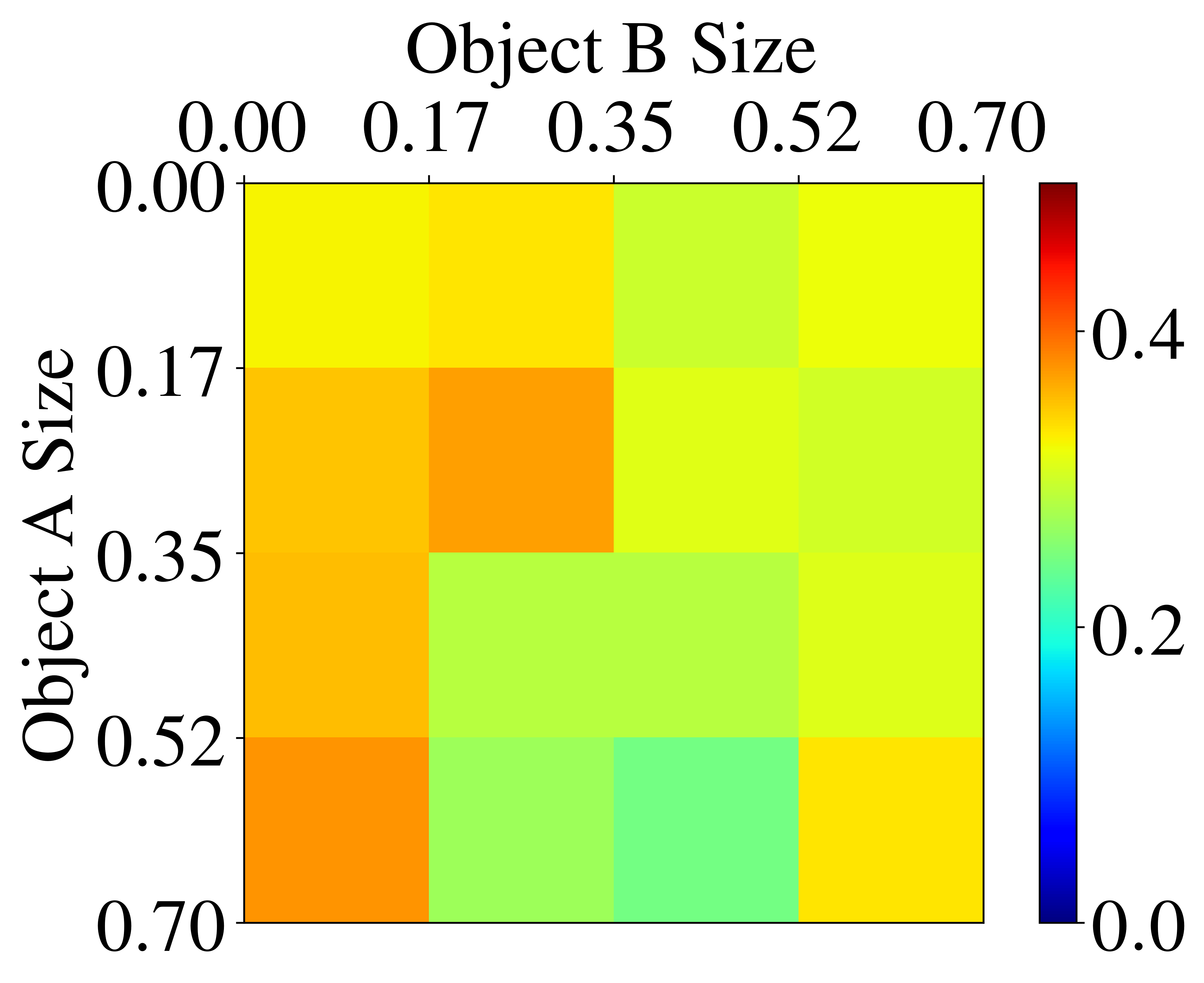}
        \vspace{-21pt}
        \caption{Across-image}
        \label{fig:acrossimage_objectsize}
    \end{subfigure}
    \\
    \begin{subfigure}[t]{0.48\linewidth}
        \centering
        \includegraphics[width=\linewidth]{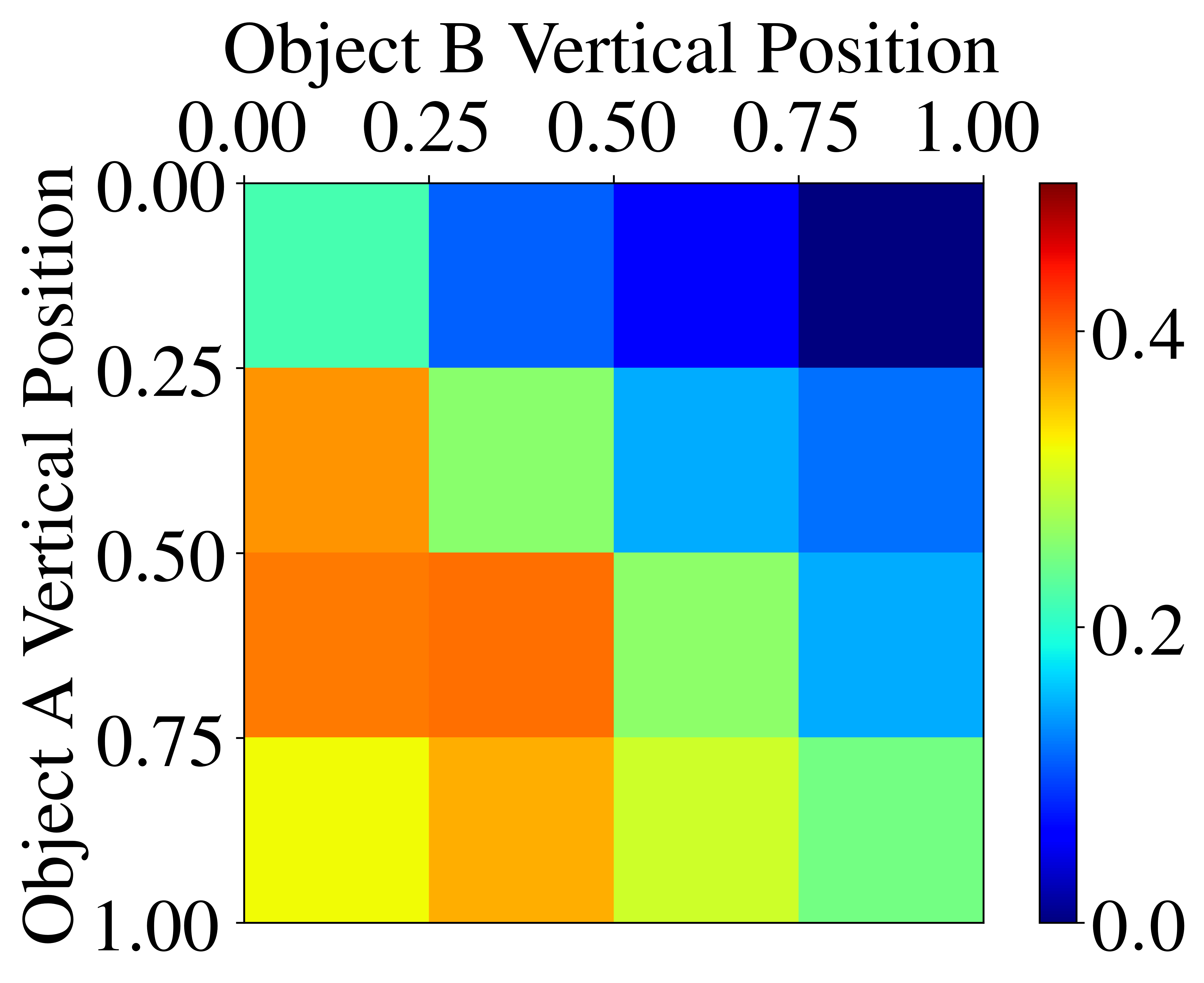}
        \vspace{-21pt}
        \caption{Within-image}
        \label{fig:withinimage_objectsize2}
    \end{subfigure}
    \begin{subfigure}[t]{0.48\linewidth}
        \centering
        \includegraphics[width=\linewidth]{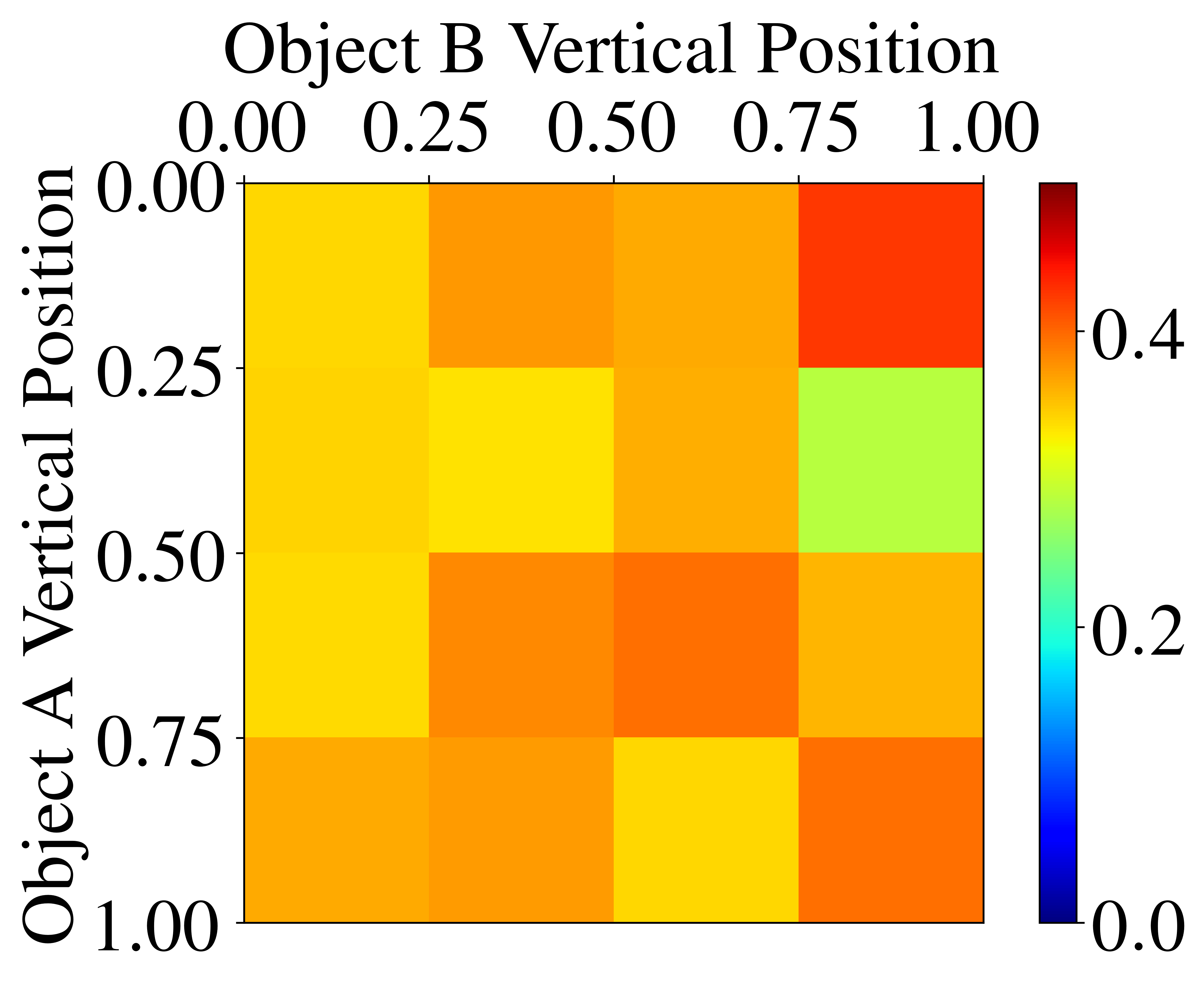}
        \vspace{-21pt}
        \caption{Across-image}
        \label{fig:acrossimage_objectsize2}
    \end{subfigure}
    \vspace{-9pt}
    \caption{
        Model performance w.r.t.\ object size and location for the $\langle o_a, \text{\em is closer than}, o_b\rangle$ relationship.
    }
    \label{fig:bbox_accuracy}
    \vspace{-12pt}
\end{figure}

\paragraph{Overall results.}
The top part of Table~\ref{tab:accuracy} shows the results of different \textbf{rule-based} methods. The estimated depth performs best on all the three \ourtask sub-tasks (within-image depth, occlusion, and across-image depth). The object location-based rule is also strong, and especially useful for the sub-task of within-image depth. In contrast, the object size matters more in the across-image depth sub-task. Finally, we see relatively low performance using the object classes. This suggests that \ourtask is more dependent on geometry than the semantic class prior.

The results of our \textbf{MLP} baseline model are in the middle part of Table~\ref{tab:accuracy}.
We explore using multiple combinations of visual cues, starting from the bounding box (B) feature only and then adding object class (C), depth (D), and appearance (A) features. We again observe that the estimated depth is a strong cue but others are also useful, with the appearance feature being most complementary, especially in the occlusion and across-image depth sub-tasks, implying the potential of the appearance cue for future work.

Since the bounding box (B) feature couples the object location and size cues, we use Figure~\ref{fig:bbox_accuracy} to dive deep into the results. We discretize the objects' (normalized) sizes and vertical positions and focus on the $\langle o_a, \text{\em is closer than}, o_b \rangle$ relations. There are high F1-scores in panels (a) and (c) when $o_a$ is larger than $o_b$ in size or the vertical position, implying that the model heeds the geometric prior for within-image \ourtask. For the same reason, the F1-scores are low when $o_a$ is closer than $o_b$ and yet $o_a$ is smaller than $o_b$ in size (or vertical position). There is no obvious pattern for the cross-image depth relations (see (b) and (d)).

The bottom part of Table~\ref{tab:accuracy} shows the results of \textbf{state-of-the-art VRD} models.
On the within-image sub-task, these sophisticated models perform comparably to our baselines.
However, they are either inapplicable or perform worse than ours in the across-image setting.
Besides, the differences between these methods are subtle. These results highlight the fact that existing VRD models cannot capture the geometry-oriented relationships in \ourtask as well as in general VRD tasks.

\begin{figure}[t]
    \centering
    \includegraphics[width=0.75\linewidth]{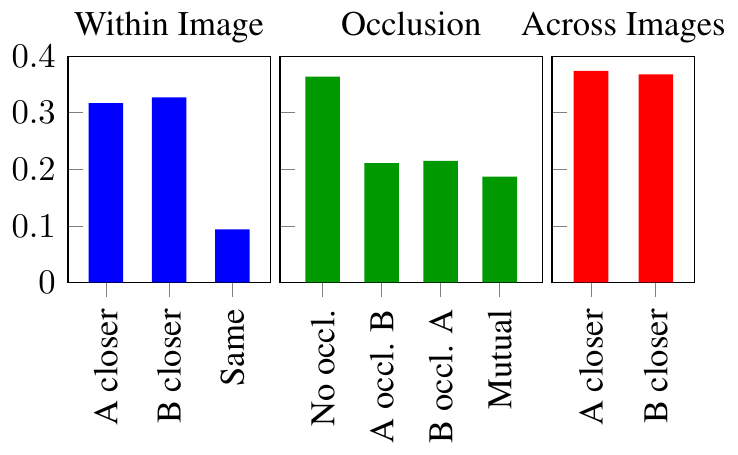}
    \vspace{-12pt}
    \caption{
        Class-wise results of the baseline model.
    }
    \label{fig:class_accuracy}
    \vspace{-15pt}
\end{figure}

\paragraph{Class-wise results.} Figure~\ref{fig:class_accuracy} categorizes the results of the baseline model (with B+C+D+A features) into different $predicates$ in each sub-task. We see that the relationship $\langle o_a, \text{\em is at the same depth as}, o_b \rangle$ is the most challenging among the within-image 2.5D relationships, probably because it happens less frequently  in the real world and in our training set (see Table~\ref{tab:stats}). The model's performance on occlusion is the lowest. 

\paragraph{Model consistency.} It is interesting to note in Figure~\ref{fig:class_accuracy} that the model's results on $\langle o_a, \text{\em is closer than}, o_b\rangle$ and $\langle o_b, \text{\em is closer than}, o_a\rangle$ are different, indicating that the model is not symmetric although we have augmented the training data by swapping all pairs of objects. We can analyze how the model meets the symmetric property more formally. Denote by $<$ and $=$ the $predicates$ of ``\emph{is closer than}'' and ``\emph{is at the same depth as}'', respectively. If the model predicts $o_a{\le}o_b$, then it is supposed to return $o_b{\ge}o_a$. We examine all pairs of objects and find that the model violates the symmetric property in 8.9\% of the test cases. Similarly, we also check the model's transitive property, i.e., if it predicts $o_a{\le}o_b$ and $o_b{\le}o_c$, then it is supposed to predict $o_a{\ge}o_c$. The model fails the transitive property test in 1.7\% of cases.
For comparison, the groundtruth labels aggregated from five raters break the transitive property in only 0.5\% of all cases.
It would be interesting to design some inductive bias into the model architecture to make its prediction symmetric and transitive in future work.

\begin{table}\small
    \centering
    \caption{Sources of error in the baseline model to \ourtask.}
    \label{tab:tasks_accuracy}
    \vspace{-9pt}
    \resizebox{\columnwidth}{!}{%
    \begin{tabular}{lccc}
     & 2.5VRD & {\bf $Predicate$ Prediction} & \underline{\it Object Detection}\\
    \midrule
    Average & 0.335 & {\bf 0.782} & \underline{\it 0.492}\\
     \bottomrule
    \end{tabular}
    }
\end{table}

\paragraph{Sources of error.}
Finally, we provide two ``upper bounds'' for the baseline model, through which we hope to understand the sources of error in \ourtask. 
Our approach takes two stages to tackle \ourtask, first detecting objects and then predicting $predicates$ for all pairs of the detected objects. We investigate from which stage the final error mainly comes from by using the following approach variations: 
\begin{itemize}[leftmargin=*,label=$\bullet$,topsep=2pt]
    \setlength{\itemsep}{1pt}
    \setlength{\parskip}{1pt}
    \item {$predicate$ prediction,} which supplies the model with groundtruth bounding boxes and classes to study how the $predicate$ prediction performs,
    \item {object detection,} singling out the object detection module by assuming a perfect $predicate$ predictor, and \item {\ourtask,} which performs both object detection and $predicate$ prediction by the full model.
\end{itemize}
Table~\ref{tab:tasks_accuracy} reports the results of the three variations. Assuming perfect object detection, \ourtask degenerates to the task of $predicate$ prediction, which boosts our method's F1-score from 0.335 to 0.782. This drastic change indicates there is a big room for the object detection module to improve for tackling \ourtask. When we use an ideal $predicate$ predictor, we only need object detection for \ourtask and observe a performance increase from 0.335 to 0.492. It is clear that the object detection module is the primary source of our model's error, but both ``upper bounds'' are virtually high. Tackling \ourtask requires advancing not only object detection but also 2.5D $predicate$ prediction.

\section{Conclusion}

We introduce \ourtask, a new task for studying the relationships between objects via depth and occlusion. 
We collect a large-scale dataset with rich human annotations, through which we conduct extensive analyses to gain insights into \ourtask. Experiments reveal that \ourtask desires progress on both object detection and predicate prediction, and the latter may benefit from a model's inductive bias that satisfies symmetric and transitive properties.

\section*{Appendices} \label{sec:appendix}

\renewcommand{\thesubsection}{\Alph{subsection}}

We supplement the main text by the following materials. 
\begin{description}
\item[Appendix~\ref{appendix:dataset-construction}] provides dataset construction details.
\item[Appendix~\ref{appendix:object-distribution}] provides the distributions of object classes in our dataset.
\item[Appendix~\ref{appendix:bias-occlusion}] analyzes the dataset's potential bias in terms of the occlusion relationships.
\item[Appendix~\ref{appendix:vrd_methods}] describes the implementation details of state-of-the-art VRD methods.
\item[Appendix~\ref{appendix:more-results}] presents more results evaluated by precision, recall, and F1-score.
\item[Appendix~\ref{appendix:transferability}] studies the models' transferability between the within-image \ourtask and across-image \ourtask.
\item[Appendix~\ref{appendix:difficulty}] analyzes model performance against different difficulty scales.
\item[Appendix~\ref{appendix:results-vs-location}] analyzes model performance against the objects' locations in an image.
\item[Appendix~\ref{appendix:more-qualitative-results}] qualitatively compares the model's predictions with the groundtruth labels.
\end{description}
If not mentioned specifically, we use the MLP baseline with all features in the analysis.

\subsection{Dataset construction} \label{appendix:dataset-construction}

This section provides the extended description of the dataset construction process presented in Section~\ref{sec:data-label-collection}. In particular, we describe the data filtering process in more detail.

We randomly sample 110,894 images from the Open Image Dataset (OID) V4.
These images all have a Creative Commons Attribution license.
Most of them are scenery images from Flickr, each containing multiple objects and/or people. 
We maintain the original train/validation/test split of the images and use the annotated bounding boxes and 600 class names  in the OID images.

As the annotations in OID are over-complete (i.e., multiple bounding boxes for an object), we filter the boxes before collecting \ourtask labels.
We remove the boxes of human body parts and clothing if the person box is available. 
Next, we discard the object pairs whose two boxes are highly overlapped (intersection-over-union is greater than $0.7$). 
In addition, we exclude the pairs of one object being part of the other (e.g., auto part and vehicle).
To avoid ill-defined cases, we ignore extremely small or big boxes (occupying less than $2\%$ or more than $70\%$ area of the image).
Finally, we remove any box that contains not one object, but a group of objects using the original OID label.

\begin{figure*}
    \centering
    \includegraphics[width=1.25\textwidth,angle=90,origin=c]{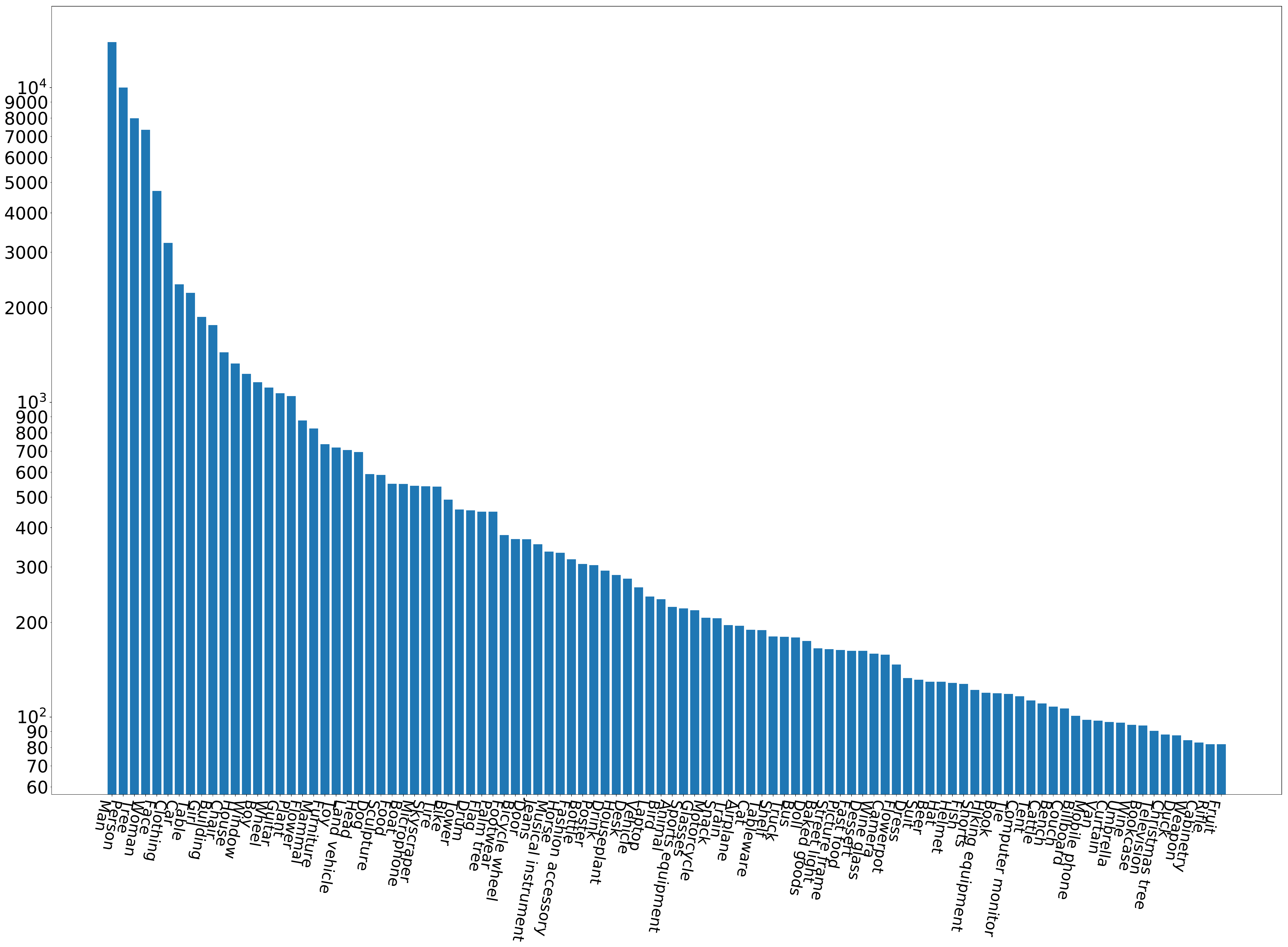}
    \caption{The distribution of top-100 object class labels, sorted by the log of frequency.}
   \label{fig:objclassdist}
\end{figure*}

\begin{figure*}
    \centering
    \includegraphics[width=1.25\textwidth,angle=90,origin=c]{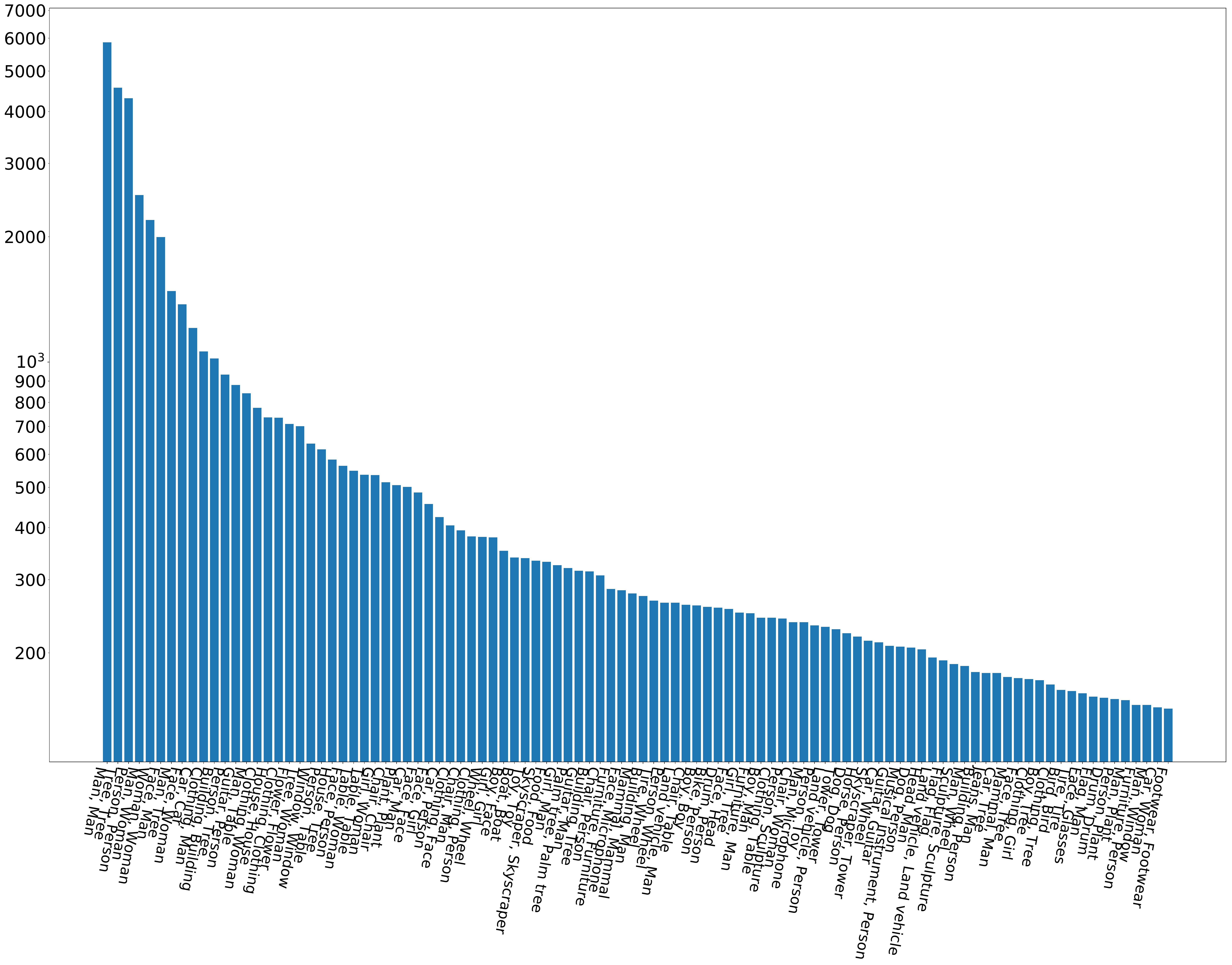}
    \caption{The distribution of top-100 pairs of object class labels, sorted by the log of frequency.}
   \label{fig:objpairclassdist}
\end{figure*}

\subsection{Object class distribution} \label{appendix:object-distribution}

Figure~\ref{fig:objclassdist} and Figure~\ref{fig:objpairclassdist} show the sizes of top-100 single object classes and the distribution of top-100 pairs of object classes, respectively. See Section~\ref{sec:data-label-collection} on how we process bounding boxes of object classes to arrive at these distributions. The most frequent object classes and pairs are human-centric --- they are about people or the objects with which people interact the most, indicating that the dataset is a fair representation of our daily scenes.

\begin{figure*}
    \centering
    \includegraphics[width=0.49\textwidth]{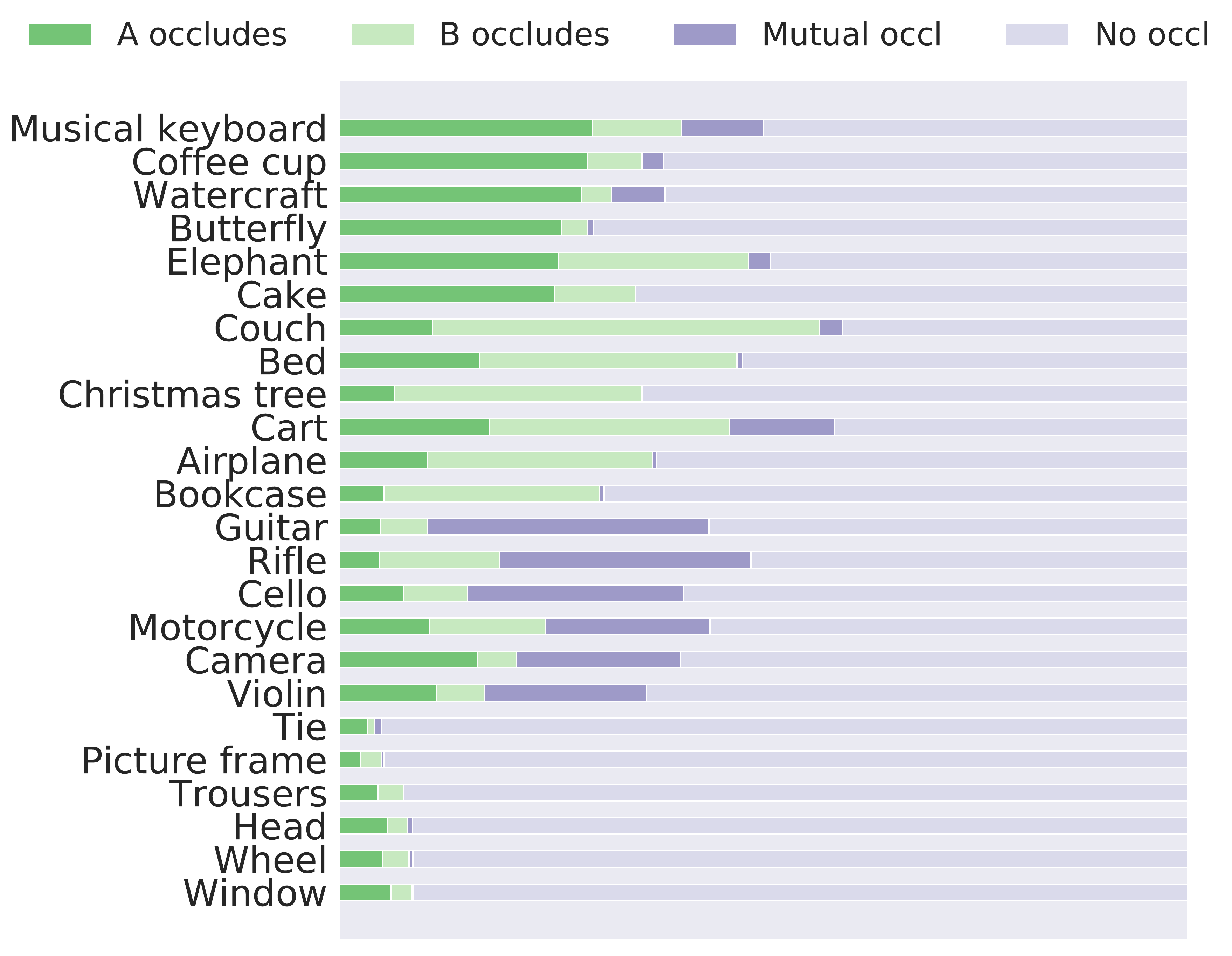}
    \includegraphics[width=0.49\textwidth]{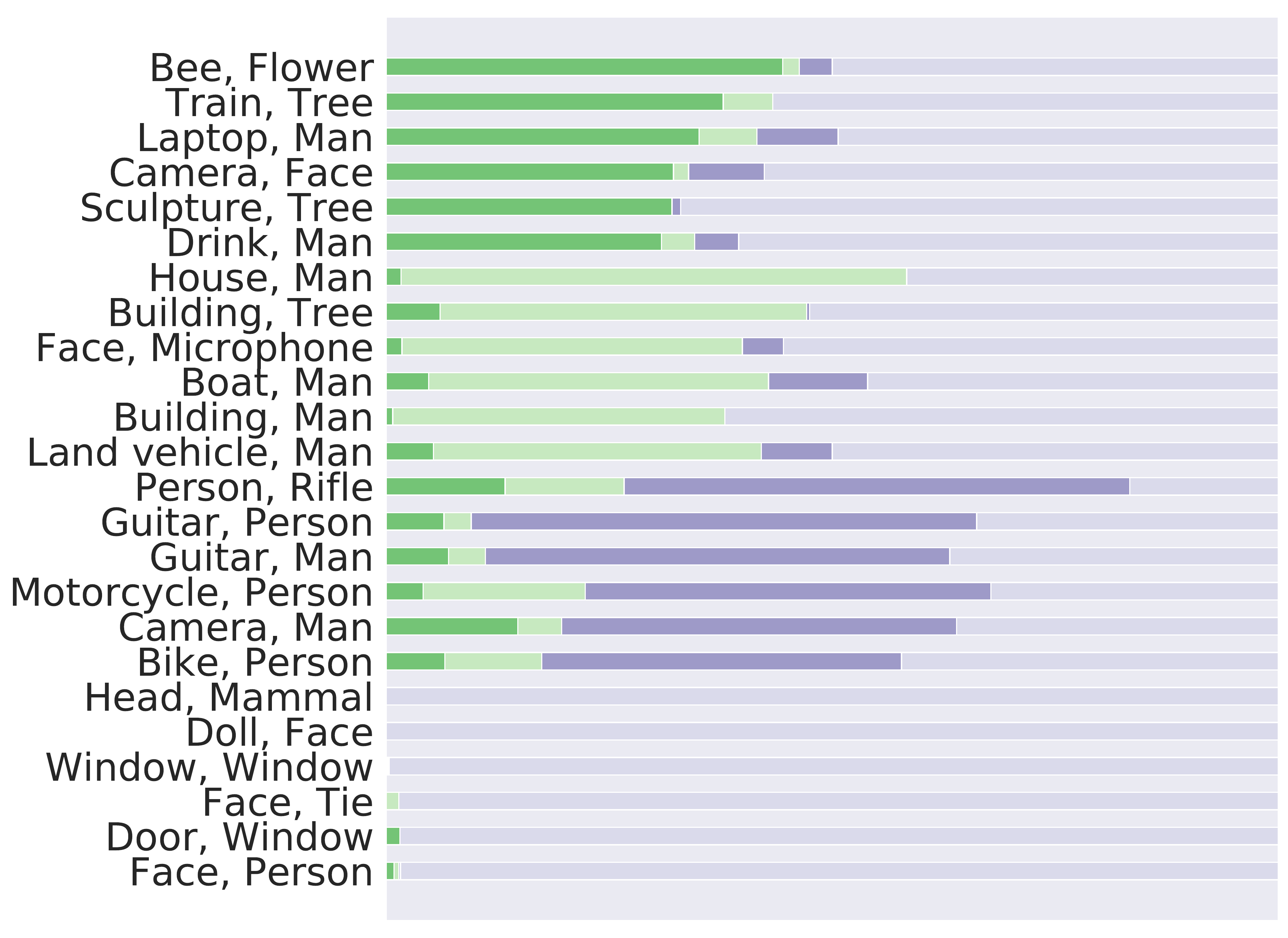}
    \vspace{-9pt}
    \caption{Distributions of occlusion labels with respect to object classes (Y-axis: Object A) and object pairs (Y-axis: (Object A, Object B)), respectively.}
    \label{fig:occlusiondist}
\end{figure*}

\subsection{Potential bias of occlusion relationships} \label{appendix:bias-occlusion}

In Section~\ref{sec:stats}, we discuss the dataset's potential bias of the depth relationships between objects. Similarly, we focus on the within-image scenario and investigate occlusion labels.
Figure~\ref{fig:occlusiondist} shows the top six object classes (and object pairs) with the highest percentage for each label. We also observe a bias. For instance, objects that ``interact'' with human body parts, such as musical instruments (guitar, cello), vehicles (motorcycle, bike), rifle, or camera, tend to be part of ``Mutual occlusion.'' We also observe that salient, often small objects (keyboard, laptop, camera, coffee cup, butterfly, bee, elephant, sculpture, elephant, train) tend to be occluded by less salient, bigger objects or stuff (couch, bed, tree, boat, airplane, building), or human/body parts (face, man). Finally, objects that are likely to be part of ``No occlusion'' are window, door, picture frame, face, and tie. We keep them as is, rather than correct them, as they are a result of the natural prior of the visual world and daily scenes.

\subsection{State-of-the-art VRD methods} \label{appendix:vrd_methods}

We implement state-of-the-art VRD methods based on the implementation released with the SpatialSense dataset\footnote{\url{https://github.com/princeton-vl/SpatialSense}}~\cite{yang2019spatialsense}.
For fair comparison, we use the same appearance feature as the MLP baselines, i.e., a Faster-RCNN pre-trained on OID.
For location features, we encode the object bounding boxes using binary masks following \textbf{DRNet}~\cite{dai2017detecting}.
For word vector embedding, we learn the word embedding for OID classes end-to-end without pre-training.
The appearance feature is used in all four methods, i.e., \textbf{ViP-CNN}~\cite{li2017vip}, \textbf{PPR-FCN}~\cite{zhuang2017towards}, \textbf{DRNet}~\cite{dai2017detecting}, and \textbf{VTransE}~\cite{zhang2017visual},
and the location and word vector embedding features are used in \textbf{DRNet} and \textbf{VTransE}.
All methods are trained using the same setup as the MLP baselines.
We verify our implementations on the SpatialSense dataset and achieve comparable accuracy as that reported in the SpatialSense paper (e.g., $71.3\%$ vs.~$71.0\%$ for \textbf{DRNet}).

\begin{table*}\small
    \centering
    \caption{
        Precision and recall of rule-based models (top part of the table), our approach with different visual cues (middle of the table; B: bounding box feature, C: object class feature, D: depth feature, A: appearance feature), and existing methods (bottom part of the table). 
    }
    \label{tab:precision_recall}
    \vspace{-10pt}
    \begin{tabular}{lccc|c}
                 & Within Image & Occlusion & Across Images & Average \\
    \midrule
    Rule: Object class     & 0.191 / 0.006 & 0.133 / 0.134 & 0.335 / 0.013 & 0.219 / 0.051\\
    \underline{\it Rule Location} & 0.264 / 0.312 & 0.302 / 0.304 & 0.205 / 0.225 & 0.257 / 0.280\\
    Rule: Size     & 0.214 / 0.253 & 0.302 / 0.304 & 0.229 / 0.251 & 0.248 / 0.269\\
    {\bf Rule: Depth} & 0.270 / 0.319 & 0.302 / 0.304 & 0.289 / 0.317 & 0.287 / 0.313\\
    \midrule
    MLP: B            & 0.237 / 0.226 & 0.307 / 0.309 & 0.232 / 0.255 & 0.259 / 0.263\\
    MLP: B+C          & 0.282 / 0.278 & 0.316 / 0.318 & 0.300 / 0.330 & 0.300 / 0.309\\
    MLP: B+D          & 0.293 / 0.310 & 0.306 / 0.309 & 0.312 / 0.342 & 0.304 / 0.320\\
    \underline{\it MLP: B+A} & 0.298 / 0.316 & 0.319 / 0.321 & 0.353 / 0.383 & 0.323 / 0.340\\
    {\bf MLP: B+C+D+A}      & 0.301 / 0.321 & 0.322 / 0.325 & 0.356 / 0.384 & 0.326 / 0.343\\
    \midrule
    ViP-CNN & 0.333 / 0.339 & 0.341 / 0.343 & - & -\\
    PPR-FCN & 0.330 / 0.341 & 0.338 / 0.340 & - & -\\
    DRNet   & 0.339 / 0.337 & 0.343 / 0.345 & 0.354 / 0.380 & 0.345 / 0.354\\ 
    VTransE & 0.315 / 0.332 & 0.328 / 0.330 & 0.353 / 0.379 & 0.332 / 0.347 \\
    \bottomrule
    \end{tabular}
\end{table*}

\begin{table*}\small
    \centering
    \caption{
        Predicate-wise 2.5VRD results of rule-based models (top part of the table), our approach with different visual cues (middle; B: bounding box feature, C: object class feature, D: depth feature, A: appearance feature), and existing method (bottom). 
    }
    \label{tab:full_accuracy}
    \vspace{-7pt}
    \begin{tabular}{lccccccccc}
                 & \multicolumn{3}{c}{Within Image} & \multicolumn{4}{c}{Occlusion} & \multicolumn{2}{c}{Across Images} \\
                 \cmidrule(lr){2-4} \cmidrule(lr){5-8} \cmidrule(lr){9-10}
                 & A closer & B closer & Same distance & No occl.\ & A occludes B & B occludes A & Mutual & A closer & B closer \\
    \midrule
    Rule: Object class     & 0.011 & 0.010 & 0.015 & 0.178 & 0.100 & 0.064 & 0.005 & 0.023 & 0.027 \\
    \underline{\it Rule: Location} & 0.306 & 0.305 & 0.129 & 0.371 & 0.172 & 0.171 & 0.000 & 0.225 & 0.218 \\
    Rule: Size     & 0.250 & 0.250 & 0.101 & 0.371 & 0.172 & 0.171 & 0.000 & 0.256 & 0.243 \\
    {\bf Rule: Depth} & 0.324 & 0.324 & 0.103 & 0.371 & 0.172 & 0.171 & 0.000 & 0.320 & 0.311 \\
    \midrule
    MLP: B            & 0.217 & 0.261 & 0.000 & 0.343 & 0.000 & 0.000 & 0.000 & 0.235 & 0.249 \\
    MLP: B+C          & 0.298 & 0.287 & 0.000 & 0.349 & 0.101 & 0.099 & 0.012 & 0.319 & 0.310 \\
    MLP: B+D          & 0.326 & 0.306 & 0.000 & 0.343 & 0.001 & 0.001 & 0.000 & 0.328 & 0.325 \\
    \underline{\it MLP: B+A} & 0.320 & 0.316 & 0.093 & 0.364 & 0.203 & 0.199 & 0.186 & 0.371 & 0.364 \\
    {\bf MLP: B+C+D+A}      & 0.317 & 0.327 & 0.094 & 0.364 & 0.211 & 0.215 & 0.187 & 0.372 & 0.368 \\
    \midrule
    ViP-CNN & 0.343 & 0.341 & 0.246 & 0.360 & 0.239 & 0.245 & 0.339 & - & - \\
    PPR-FCN & 0.344 & 0.344 & 0.208 & 0.369 & 0.228 & 0.224 & 0.266 & - & - \\
    DRNet   & 0.342 & 0.356 & 0.175 & 0.365 & 0.259 & 0.274 & 0.190 & 0.369 & 0.364 \\
    VTransE & 0.332 & 0.335 & 0.199 & 0.352 & 0.224 & 0.228 & 0.266 & 0.367 & 0.364 \\
    \bottomrule
    \end{tabular}
\end{table*}

\subsection{Overall results} \label{appendix:more-results}

In this section, we show additional results expanded from Table~\ref{tab:accuracy}.
We first show the corresponding precision and recall in Table~\ref{tab:precision_recall}.
The results show that combining multiple features improves both precision and recall consistently.
Next, we show the F1-score for each predicate in Table~\ref{tab:full_accuracy}.
The results show that the models are not fully symmetric,
i.e.,~$o_{a}$ is closer than $o_{b}$ does not always imply $o_{b}$ is further than $o_{a}$.
Also, we can see that the appearance feature is important for occlusion prediction,
especially for mutually occluded cases.

\begin{table}\small
    \centering
    \caption{Model transferability across \ourtask sub-tasks.}
    \label{tab:transferability}
    \vspace{-7pt}
    \begin{tabular}{lcc}
        $\longrightarrow$     & Within-Image & Across-Image \\
        \midrule
        Within-Image & 0.322 & 0.308\\
        Across-Image & 0.304 & 0.370\\
        Joint & 0.318 & 0.354\\
        \bottomrule
    \end{tabular}
\end{table}

\subsection{Model transferability} \label{appendix:transferability}

In our previous experiments,
we train separate models for within-image and across-image 2.5VRD, though the models share exactly the same architecture.
However, we can unify them into one.
To evaluate how well a model generalizes across the two settings,
we test the models' transferability across within-image and across-image depth relationships.

Table~\ref{tab:transferability} shows the results.
Not surprisingly, the model performance degrades when it transfers from the within-image sub-task to the across-image sub-task, and vice versa. The unified model, which is trained by pooling the in-image and across-image training examples, is in between of the other models.
The results show that 2.5VRD models do not fulfill the desired property of transferability, which raises the need for further development of models and/or learning algorithms.

\begin{figure}
    \centering
    \begin{subfigure}[t]{0.48\linewidth}
        \centering
        \includegraphics[width=\linewidth]{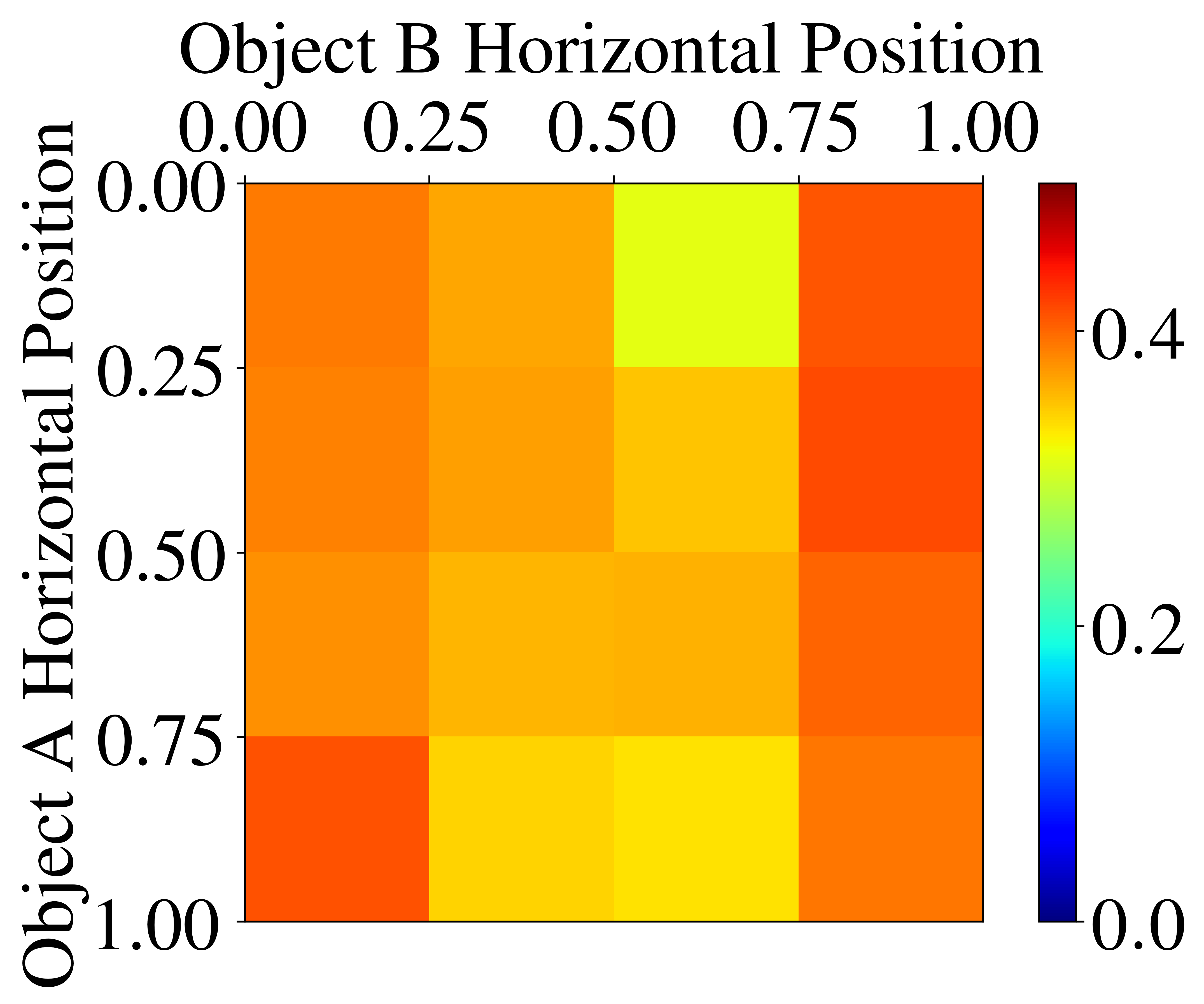}
        \vspace{-21pt}
        \caption{Within Image}
        \label{fig:withinimage_objectsize}
    \end{subfigure}
    \begin{subfigure}[t]{0.48\linewidth}
        \centering
        \includegraphics[width=\linewidth]{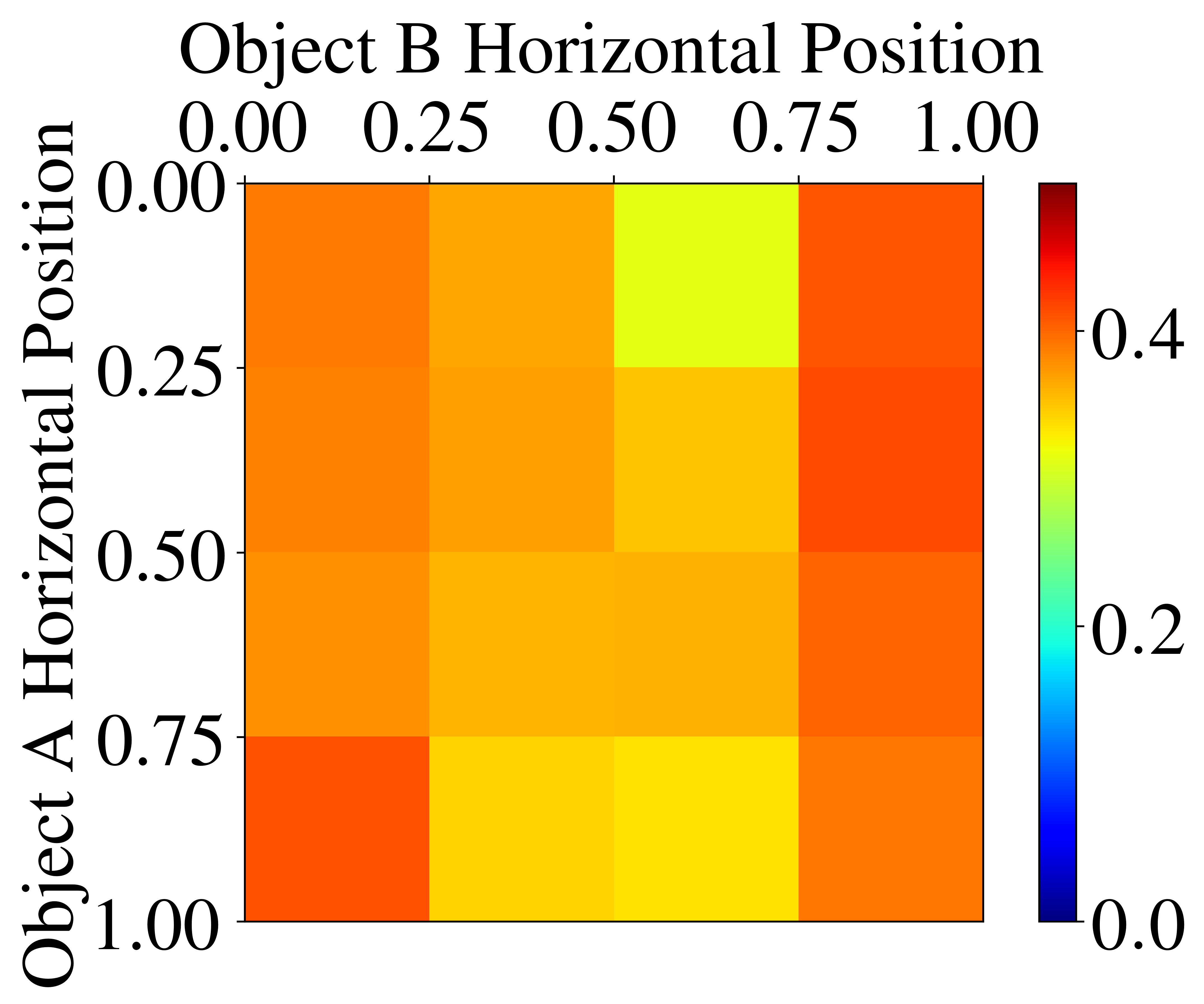}
        \vspace{-21pt}
        \caption{Across Images}
        \label{fig:withinimage_objectsize}
    \end{subfigure}
    \caption{
        Model performance w.r.t.\ objects' hotizontal locations for the $\langle o_a, \text{\em is closer than}, o_b\rangle$ relationship.
    }
    \label{fig:bbox_horizontal_accuracy}
    \vspace{-10pt}
\end{figure}

\subsection{Results of various difficulty scales} \label{appendix:difficulty}

This section shows the model performance at different difficulty scales.
The results are in Table~\ref{tab:difficulty_accuracy}.
Note that the difficulties are defined only on annotated objects, so we use the groundtruth objects in this experiment (i.e., predicate prediction in Table~\ref{tab:tasks_accuracy}).
The model's performance aligns very well with the human raters' assessments about the examples' difficulty scales. 

\begin{table}[t]\small
    \centering
    \tabcolsep=0.12cm
    \caption{2.5VRD results of various difficulty scales.}
    \label{tab:difficulty_accuracy}
    \vspace{-8pt}
    \begin{tabular}{lccc|c}
                 & Within Image & Occlusion & Across Images & Average \\
    \midrule
    {\bf Easy}                & 0.886 & 0.821 & 0.950 & 0.886\\
    \underline{\it Moderate}  & 0.644 & 0.781 & 0.833 & 0.753\\
    Difficult                 & 0.483 & 0.781 & 0.668 & 0.644\\
     \bottomrule
    \end{tabular}
\end{table}

\subsection{Object location distribution} \label{appendix:results-vs-location}

Figure~\ref{fig:bbox_accuracy} shows that the model's accuracy correlates with the objects' vertical positions in an image.
In contrast, Figure~\ref{fig:bbox_horizontal_accuracy} shows that the model performance is not sensitive to the objects' horizontal position.
The results are consistent with our observation that an object's depth is highly correlated with the Y-coordinate of the object center.

\subsection{Qualitative results} \label{appendix:more-qualitative-results}

\begin{figure*}
    \centering
    \begin{subfigure}{\textwidth}
        \centering
        \InsertImage{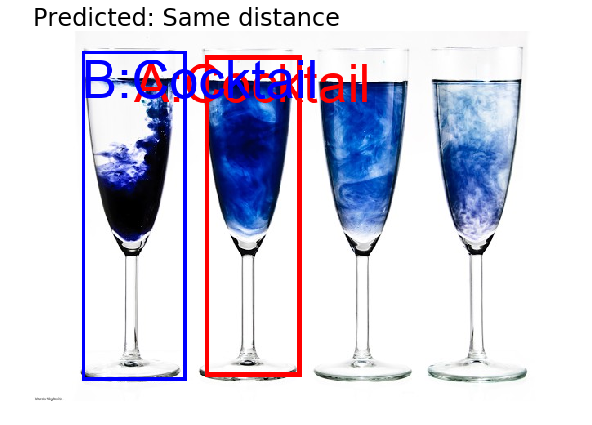}
        \InsertImage{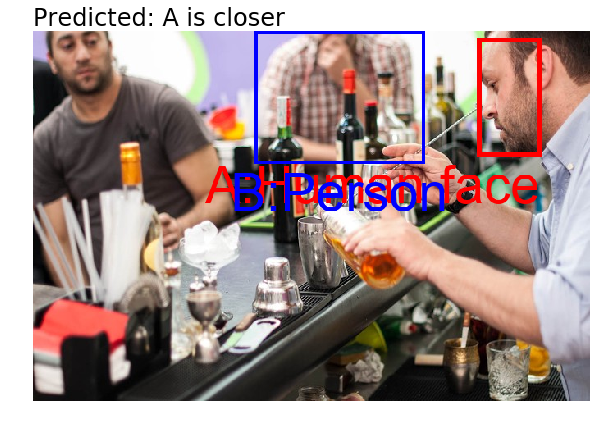}
        \InsertImage{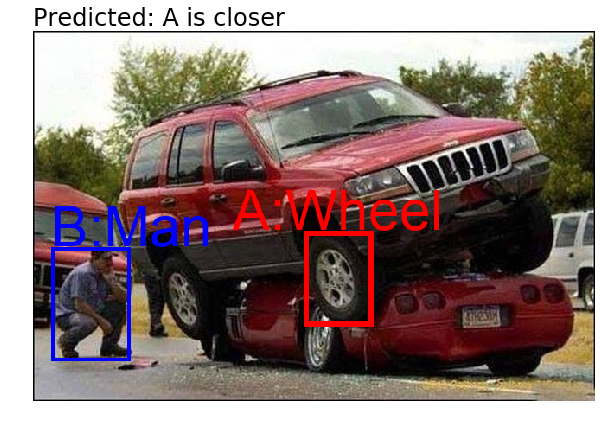}
        \InsertImage{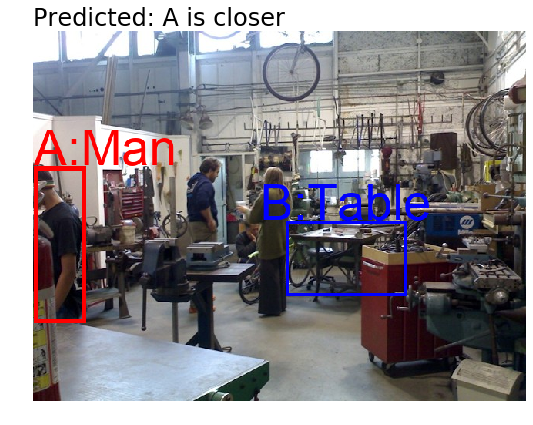}
        \\
        \InsertImage{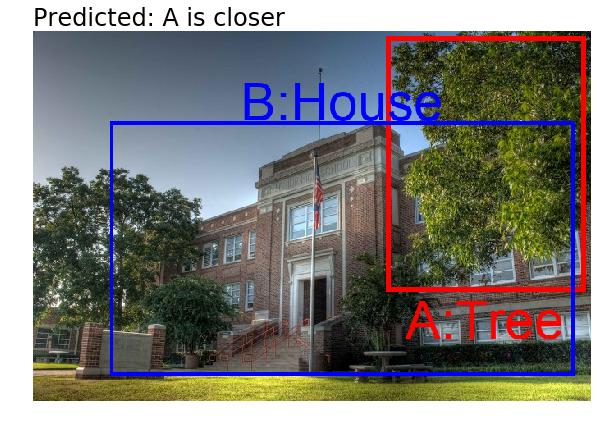}
        \InsertImage[0.15]{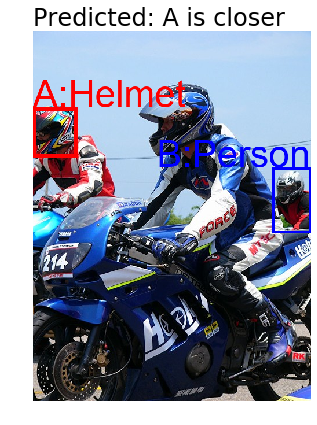}
        \InsertImage{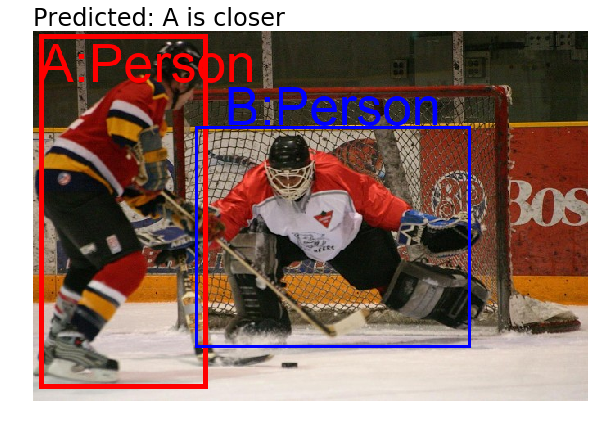}
        \InsertImage{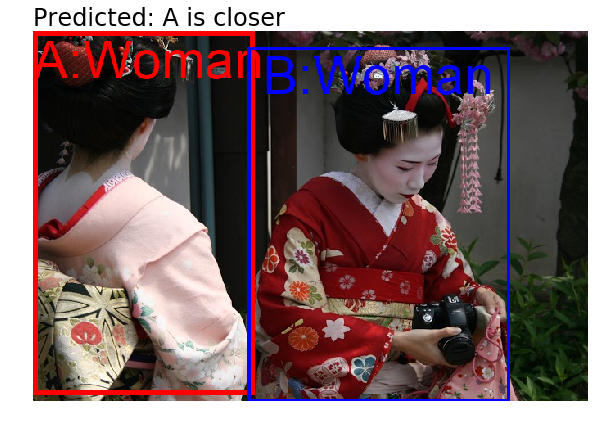}
        \vspace{-9pt}
        \caption{Easy}
        \label{fig:qualitative_easy}
    \end{subfigure}
    \begin{subfigure}{\textwidth}
        \centering
        \InsertImage[0.18]{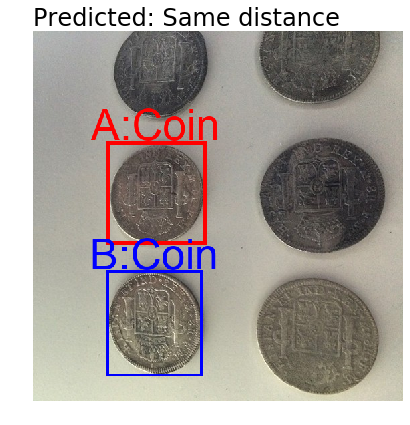}
        \InsertImage{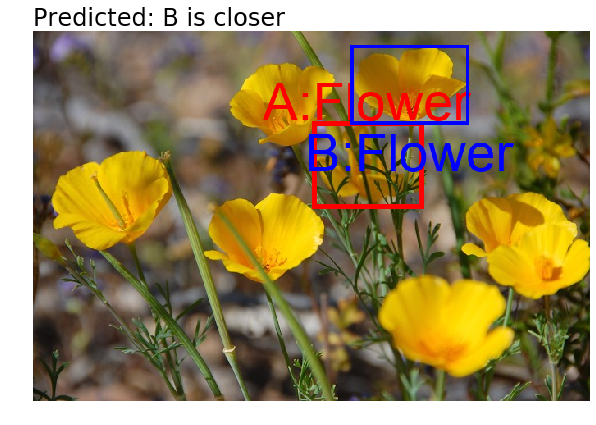}
        \InsertImage{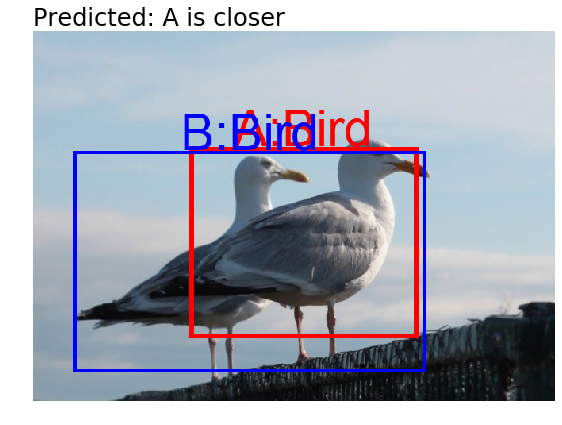}
        \InsertImage{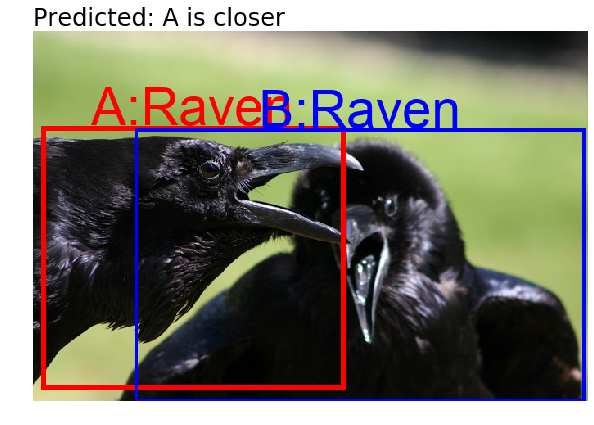}
        \\
        \InsertImage{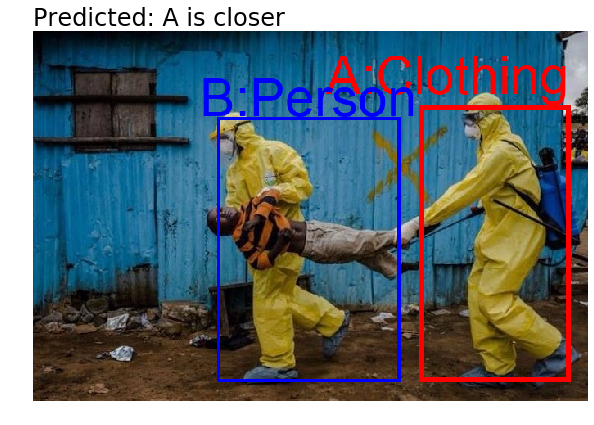}
        \InsertImage{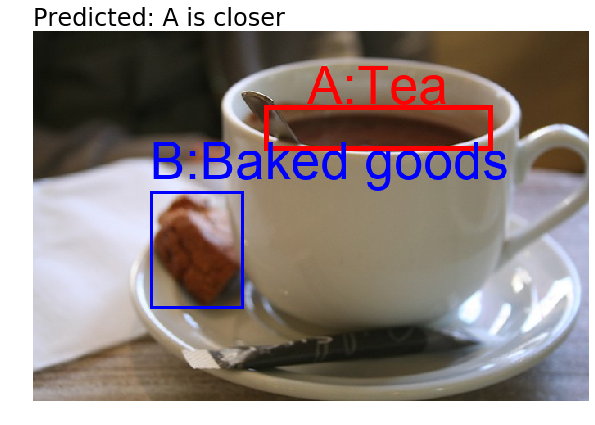}
        \InsertImage{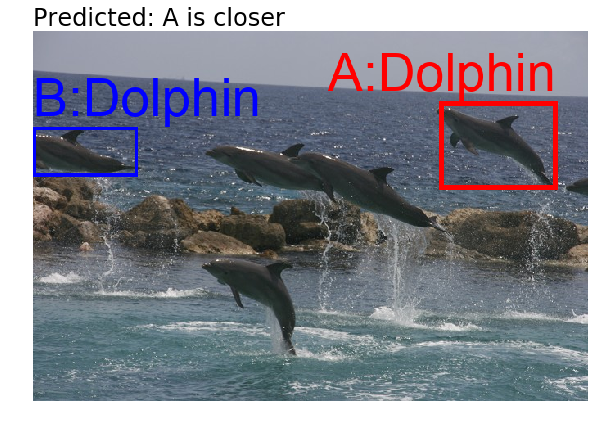}
        \InsertImage{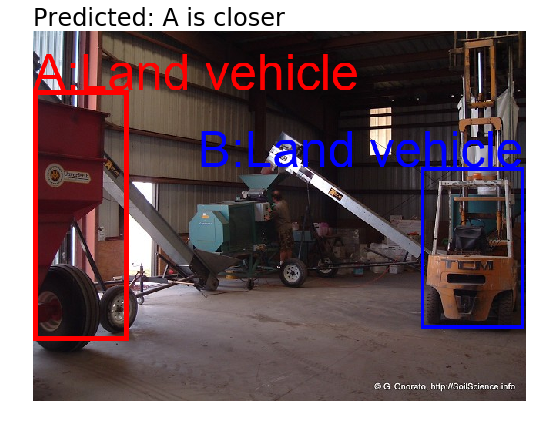}
        \vspace{-9pt}
        \caption{Moderate}
        \label{fig:qualitative_moderate}
    \end{subfigure}
    \begin{subfigure}[c]{\textwidth}
        \centering
        \InsertImage{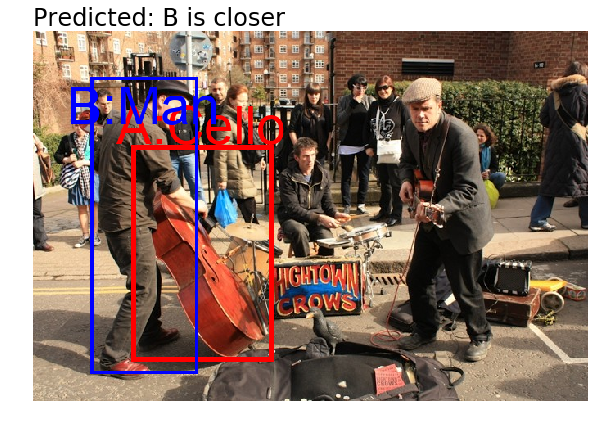}
        \InsertImage{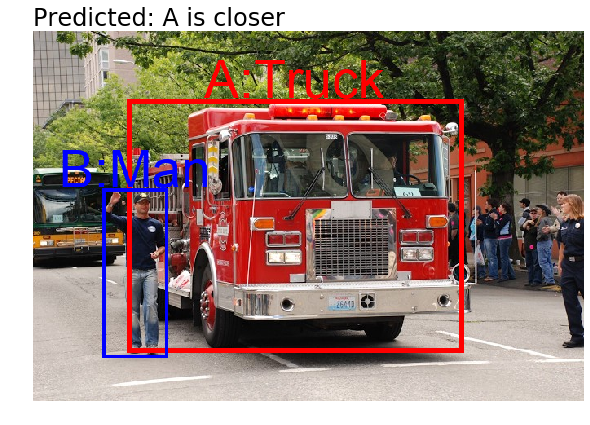}
        \InsertImage{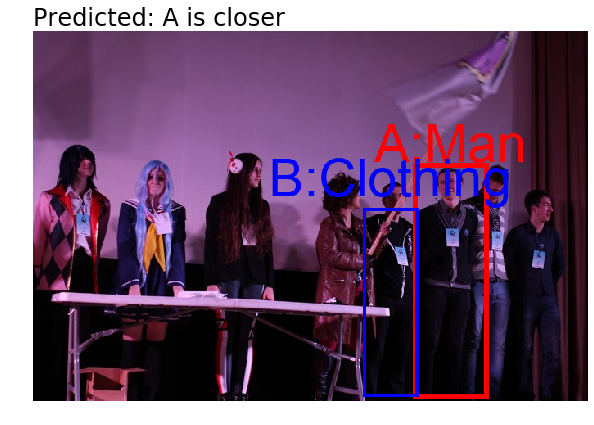}
        \InsertImage{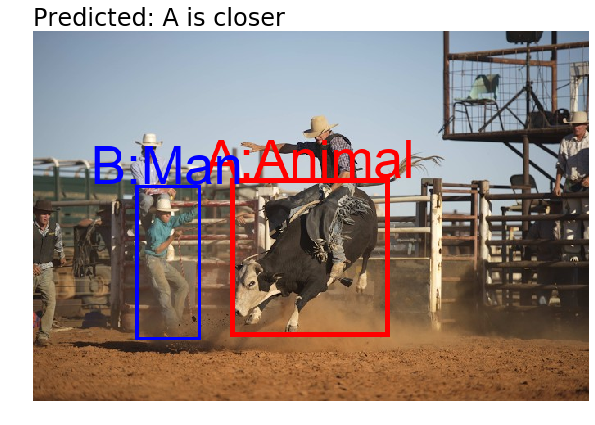}
        \\
        \InsertImage{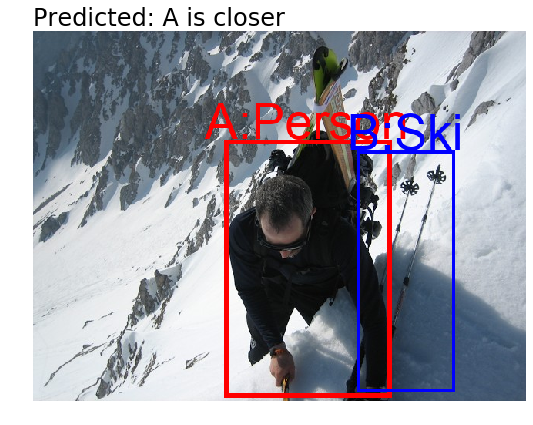}
        \InsertImage{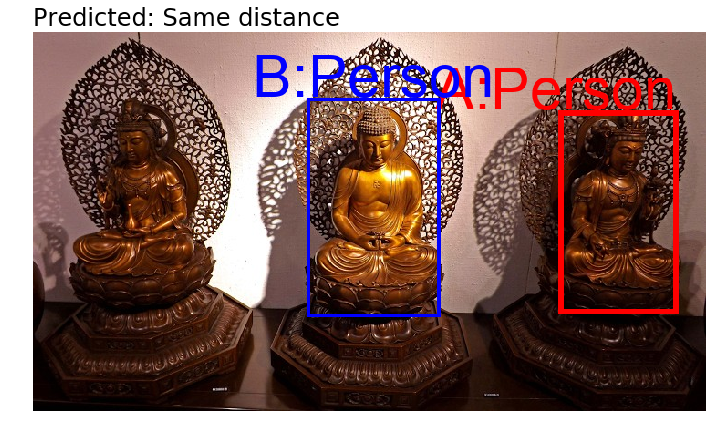}
        \InsertImage{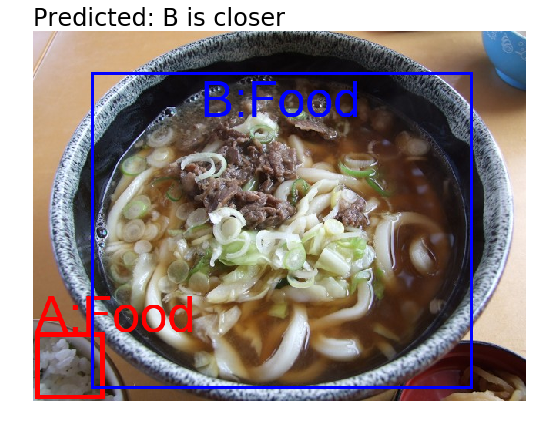}
        \InsertImage{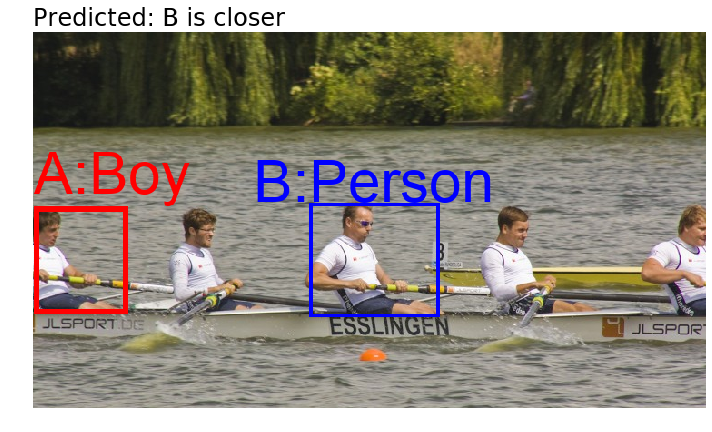}
        \vspace{-9pt}
        \caption{Hard}
        \label{fig:qualitative_hard}
    \end{subfigure}
    \caption{Qualitative examples for within-image depth prediction with different difficulties (groundtruth = predicted labels).}
    \label{fig:qualitative}
\end{figure*}

\begin{figure*}
    \centering
    \InsertImage{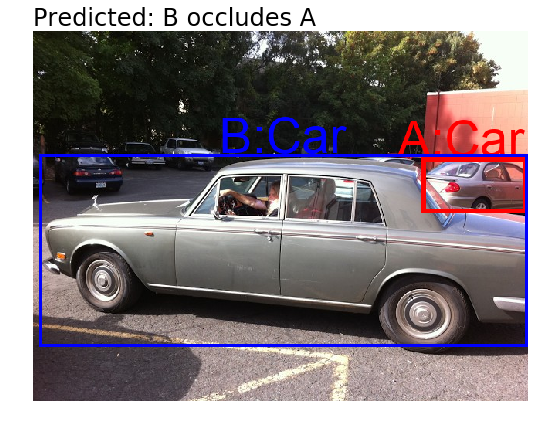}
    \InsertImage{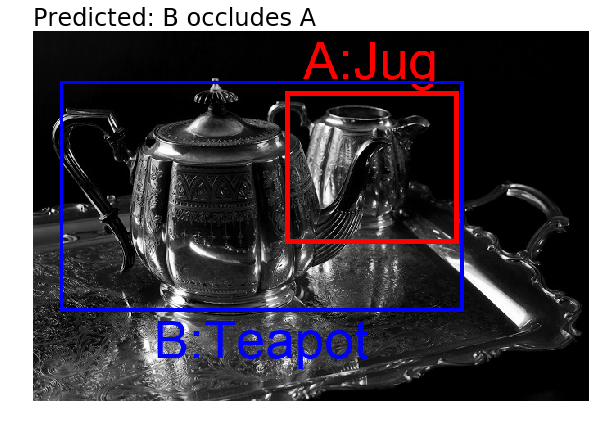}
    \InsertImage{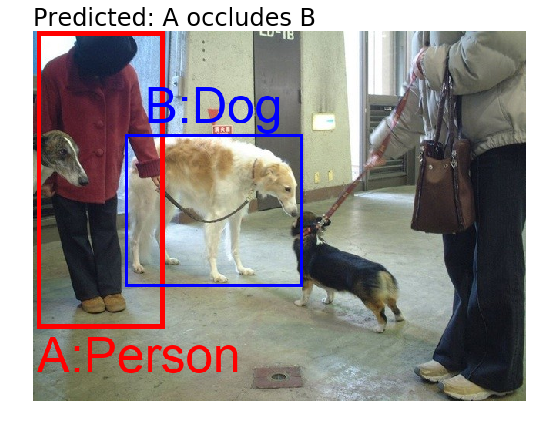}
    \InsertImage{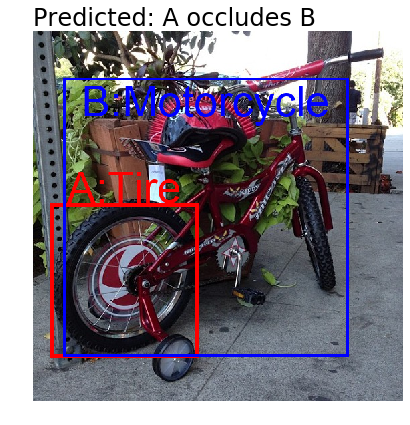}
    \\
    \InsertImage{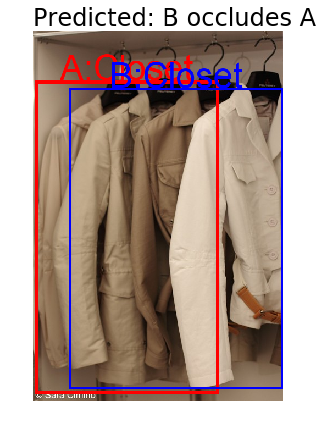}
    \InsertImage{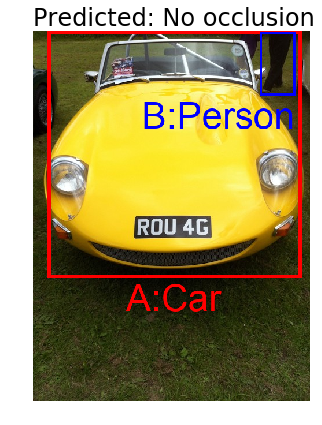}
    \InsertImage{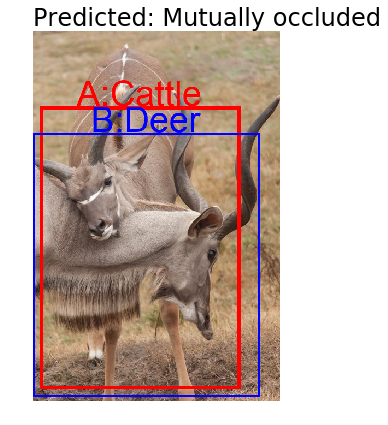}
    \InsertImage{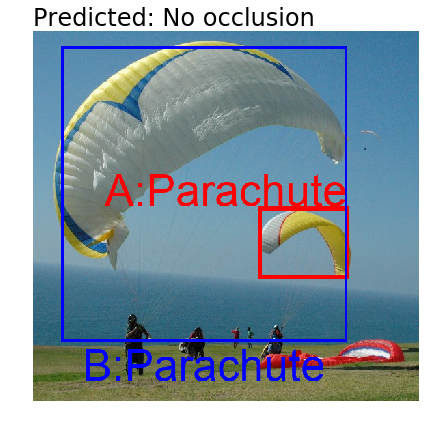}
    \\
    \InsertImage{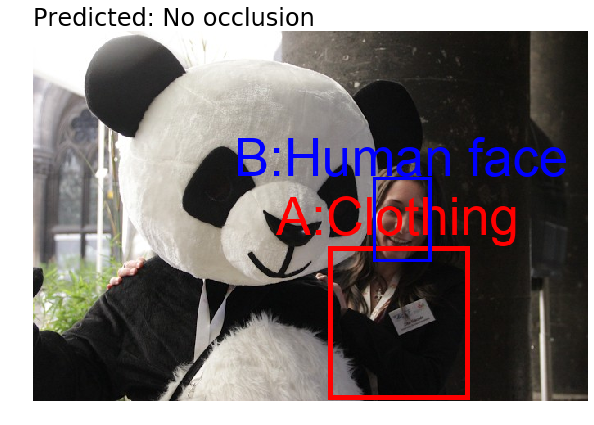}
    \InsertImage{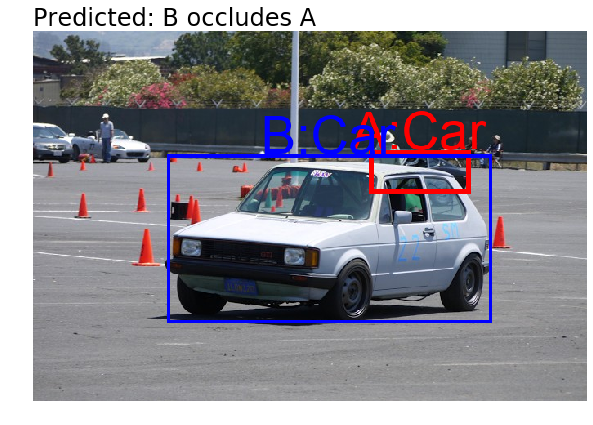}
    \InsertImage{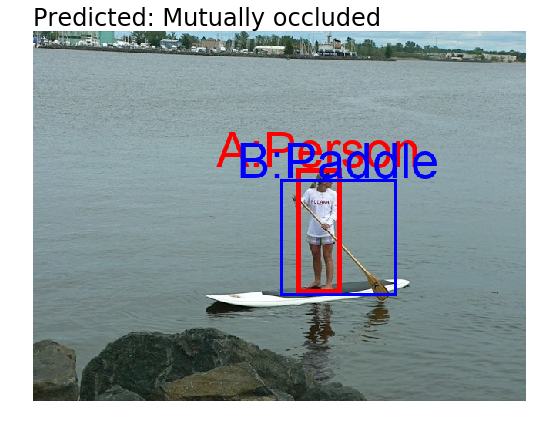}
    \InsertImage{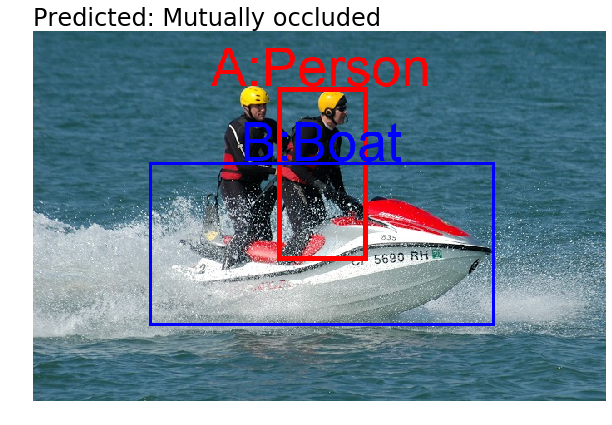}
    \\
    \InsertImage{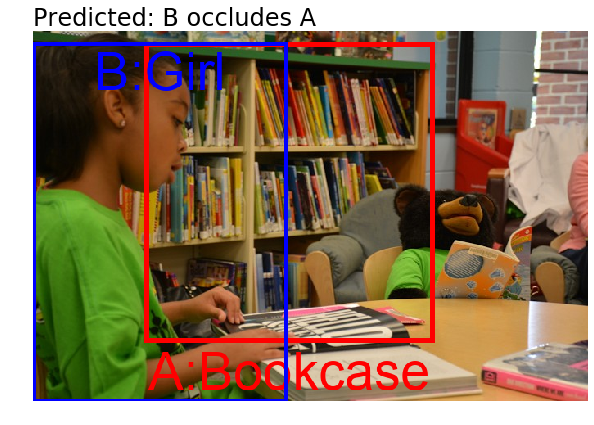}
    \InsertImage{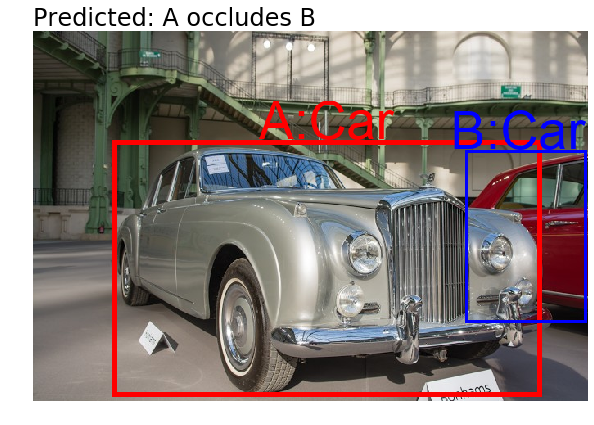}
    \InsertImage{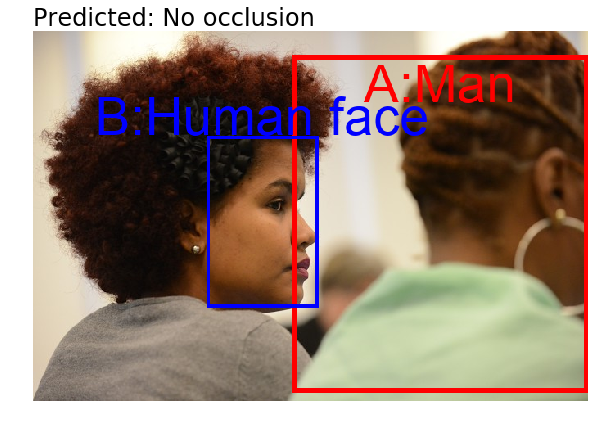}
    \InsertImage{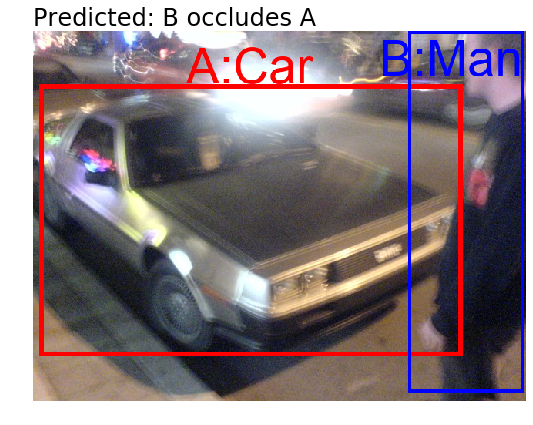}
    \\
    \InsertImage{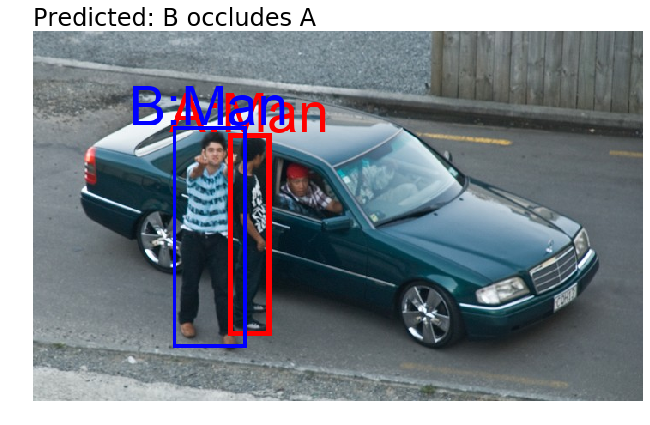}
    \InsertImage{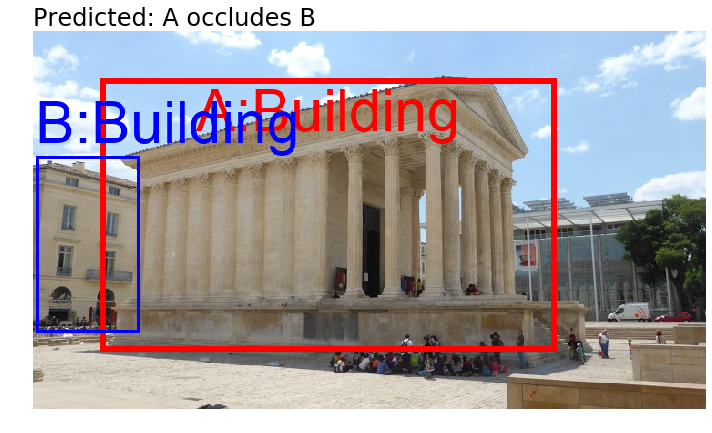}
    \InsertImage{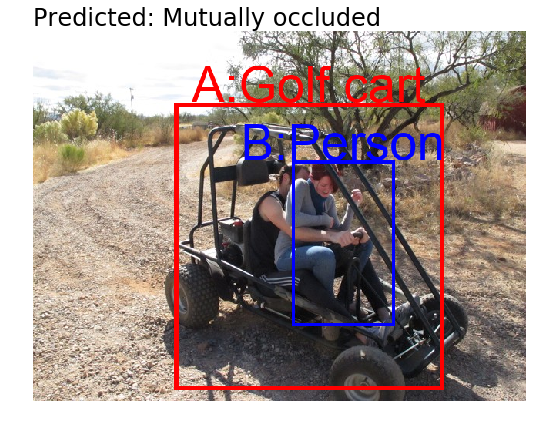}
    \InsertImage{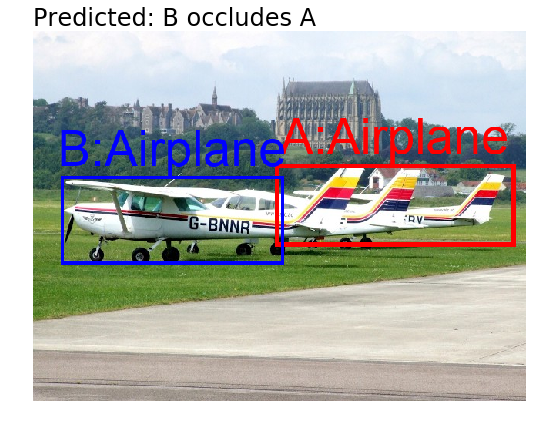}
    \\
    \caption{Qualitative examples for within-image occlusion prediction  (groundtruth = predicted labels).}
    \label{fig:qualitative_occlusion}
\end{figure*}

\begin{figure*}
    \centering
    \InsertImage[0.48]{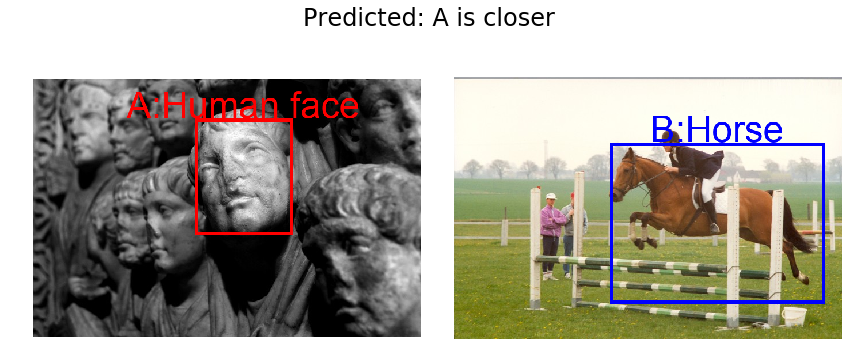}
    \InsertImage[0.48]{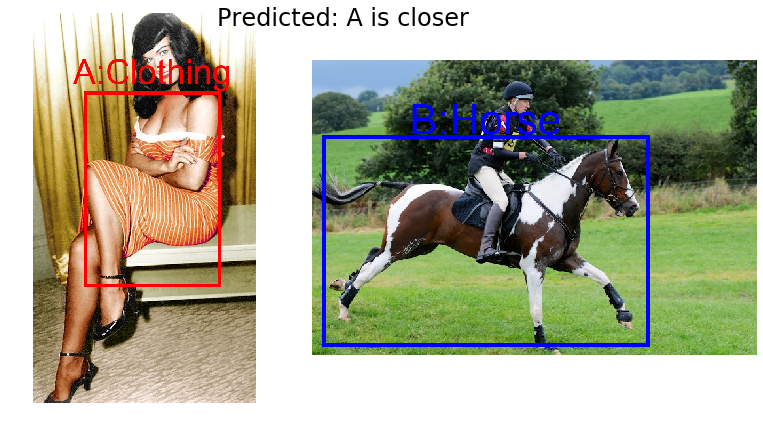}
    \\
    \InsertImage[0.48]{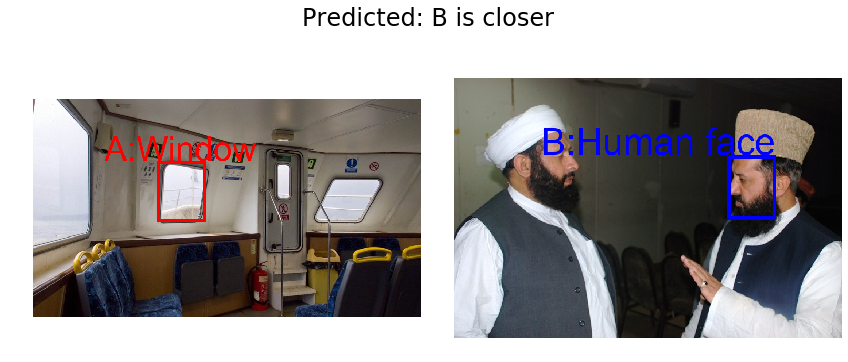}
    \InsertImage[0.48]{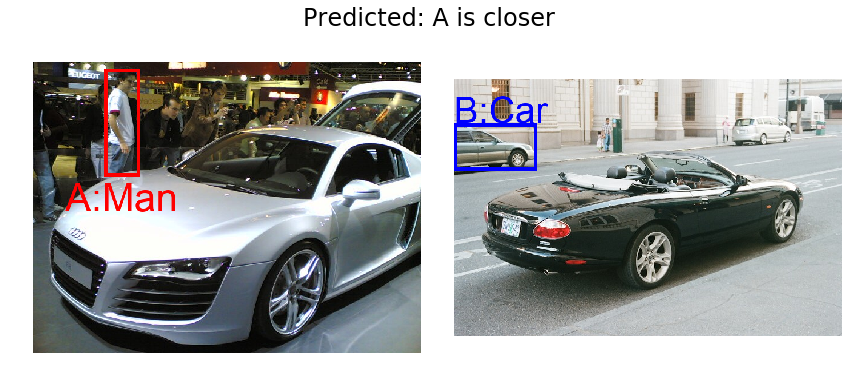}
    \\
    \InsertImage[0.48]{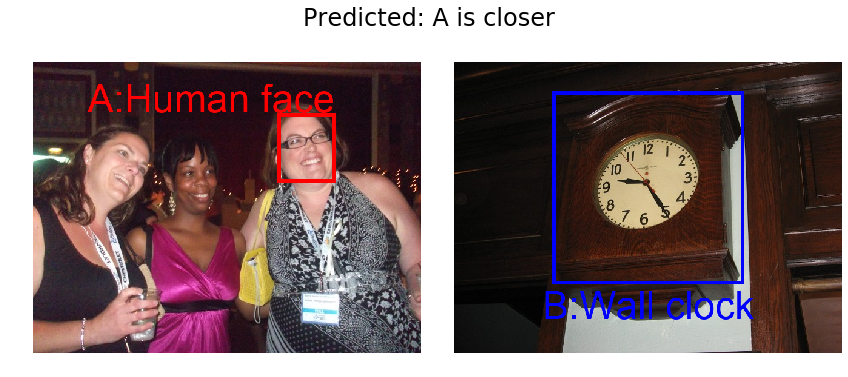}
    \InsertImage[0.48]{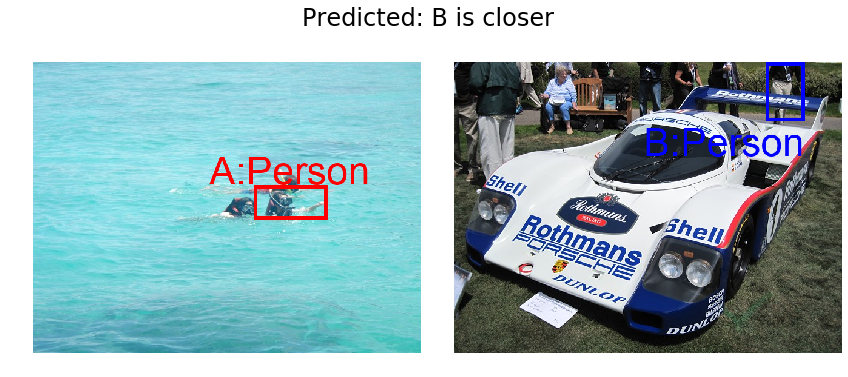}
    \\
    \InsertImage[0.48]{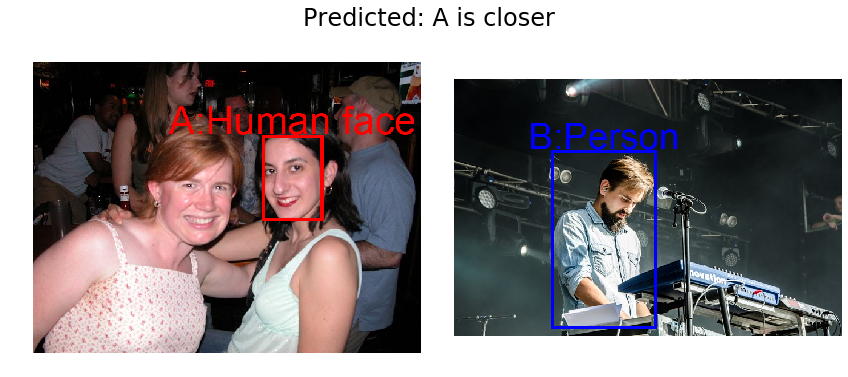}
    \InsertImage[0.48]{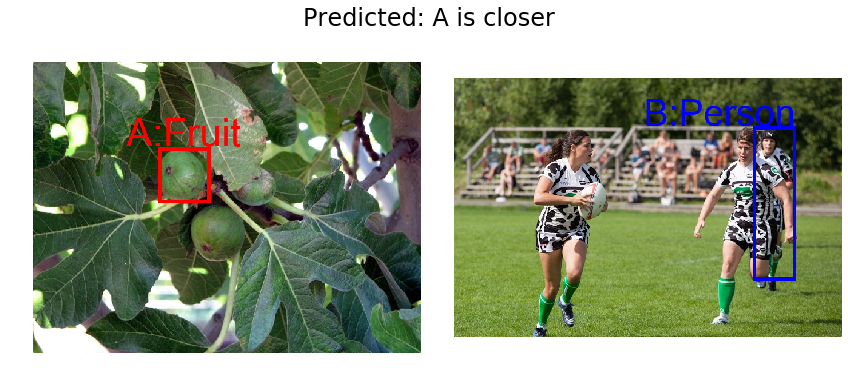}
    \\
    \InsertImage[0.48]{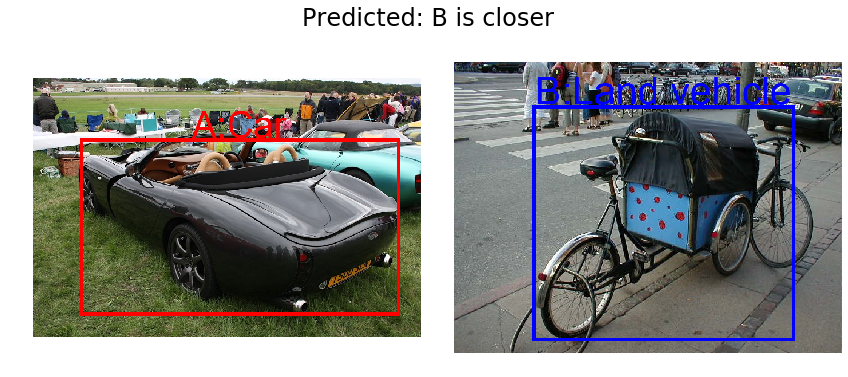}
    \InsertImage[0.48]{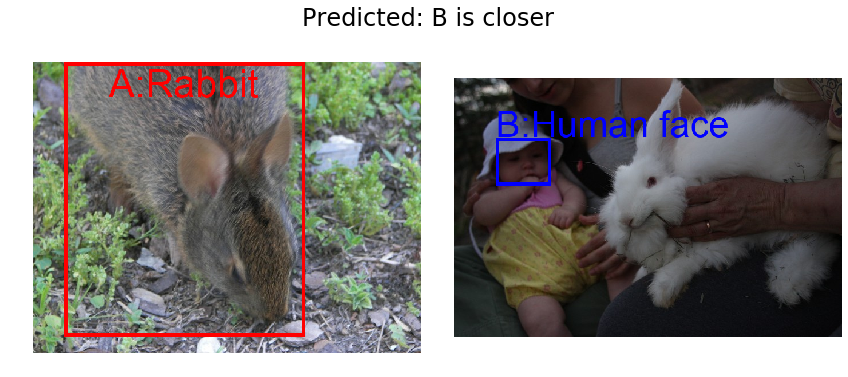}
    \\
    \caption{Qualitative examples for across-image depth prediction  (groundtruth = predicted labels).}
    \label{fig:qualitative_across}
\end{figure*}

\begin{figure*}
    \centering
    \InsertImage{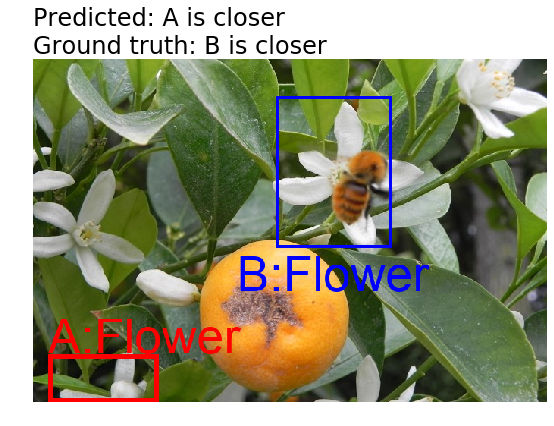}
    \InsertImage{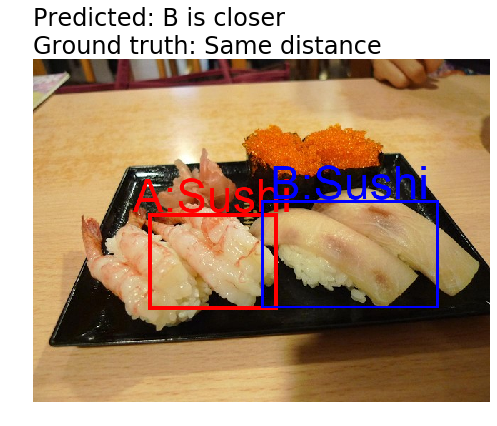}
    \InsertImage{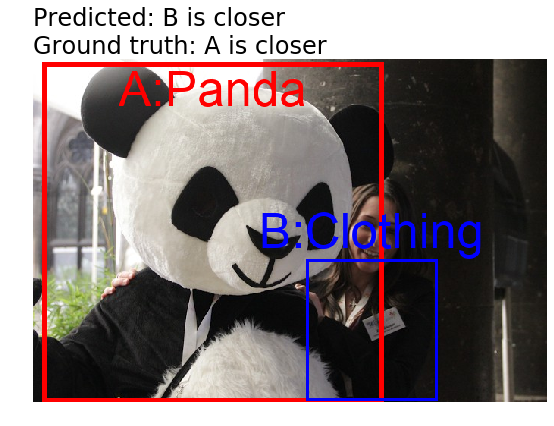}
    \InsertImage{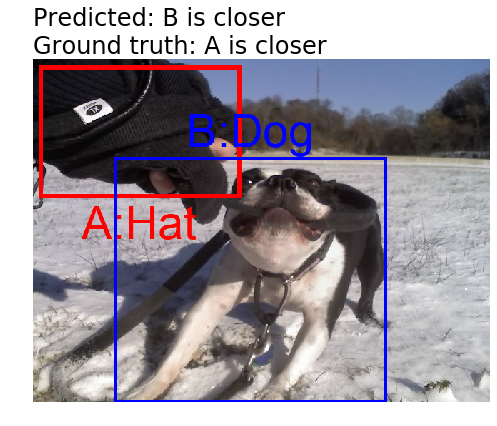}
    \\
    \InsertImage{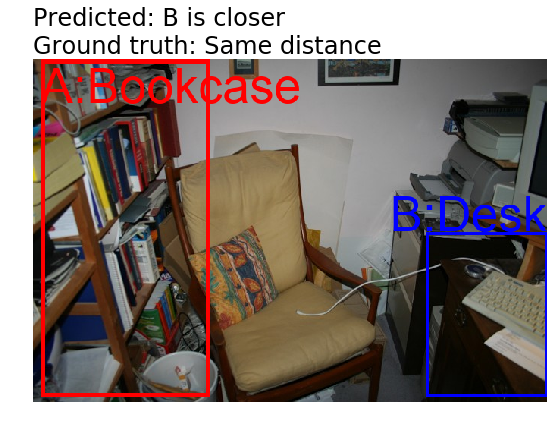}
    \InsertImage{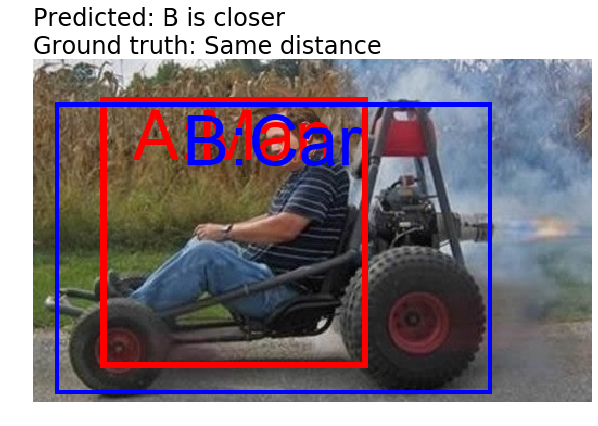}
    \InsertImage{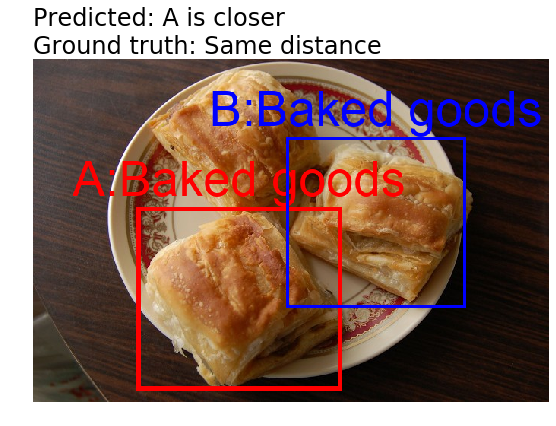}
    \InsertImage{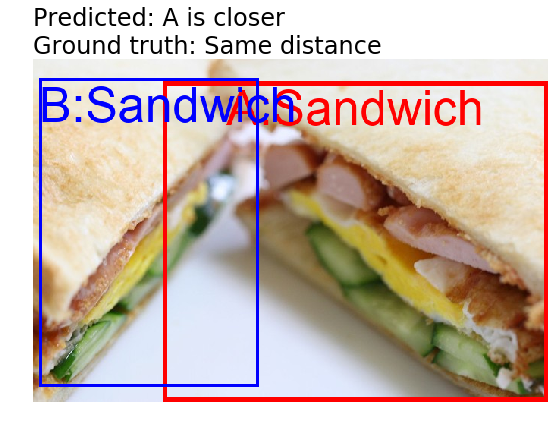}
    \\
    \InsertImage[0.2]{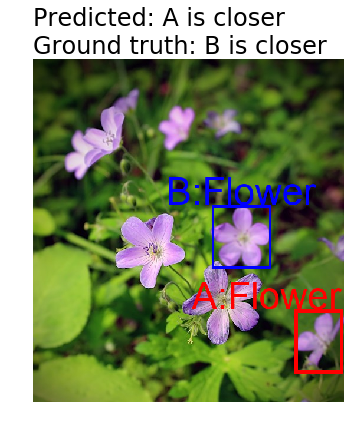}
    \InsertImage{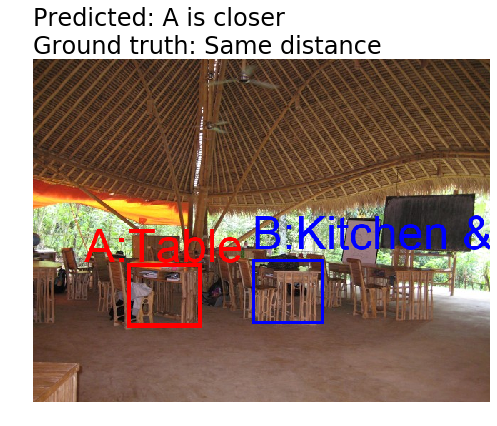}
    \InsertImage{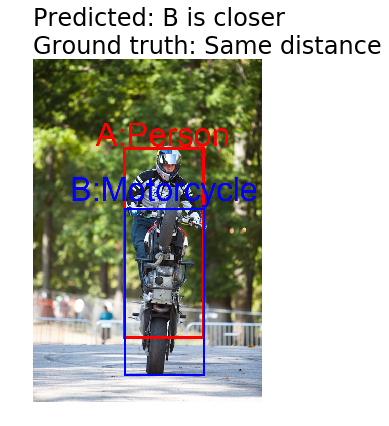}
    \InsertImage{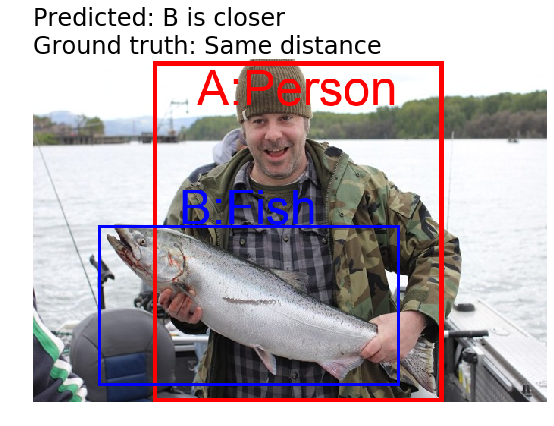}
    \\
    \caption{Failure cases for within-image depth prediction.
    This figure show examples where the model correctly detects the objects but predicts wrong predicates.}
    \label{fig:failure_relation}
\end{figure*}

\begin{figure*}
    \centering
    \InsertImage{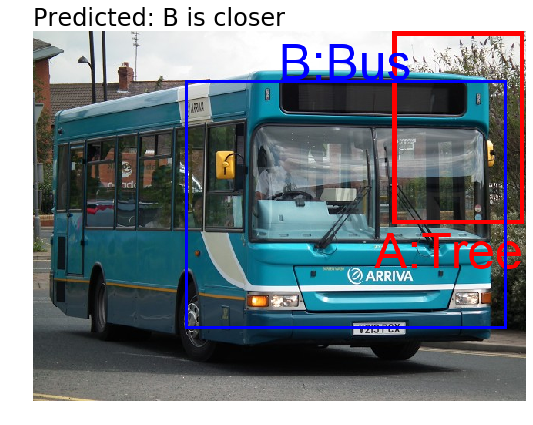}
    \InsertImage{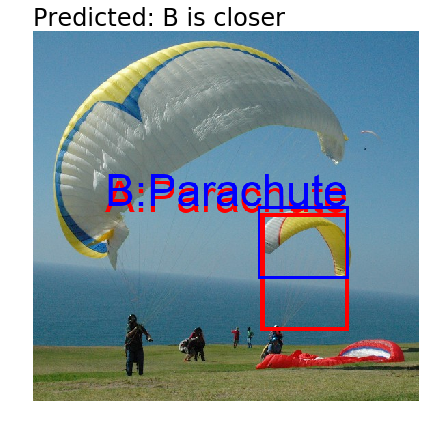}
    \InsertImage[0.18]{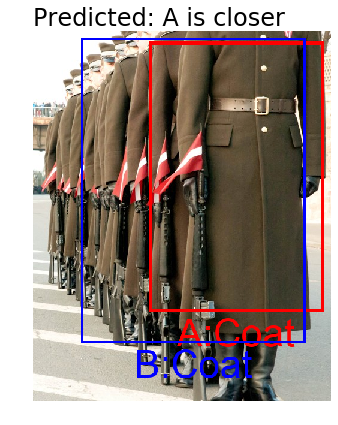}
    \InsertImage{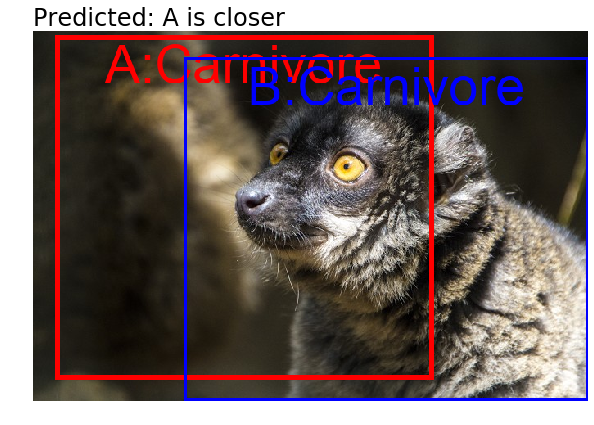}
    \\
    \InsertImage{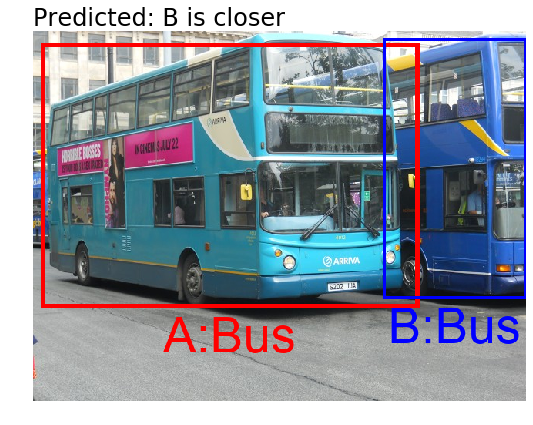}
    \InsertImage{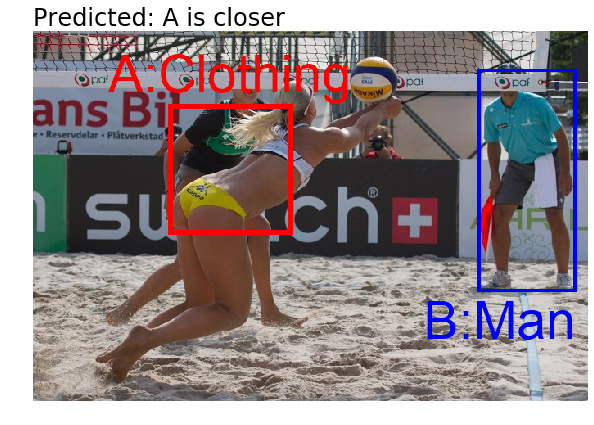}
    \InsertImage[0.18]{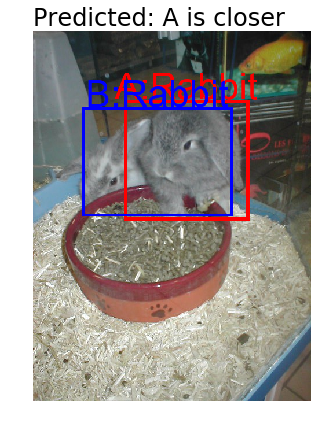}
    \InsertImage{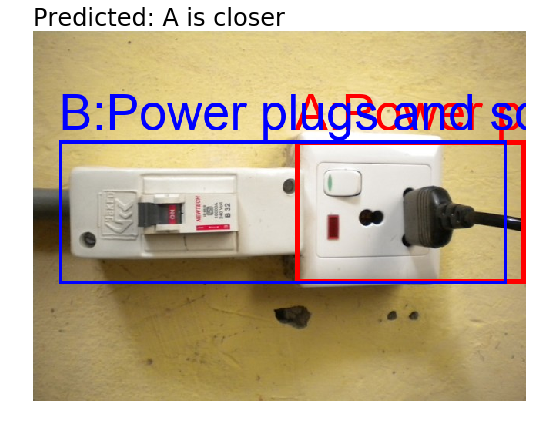}
    \\
    \caption{Failure cases for within-image depth prediction. This figure show examples where the model fails to detect the objects.}
    \label{fig:failure_detection}
\end{figure*}

\begin{figure*}
    \centering
    \InsertImage{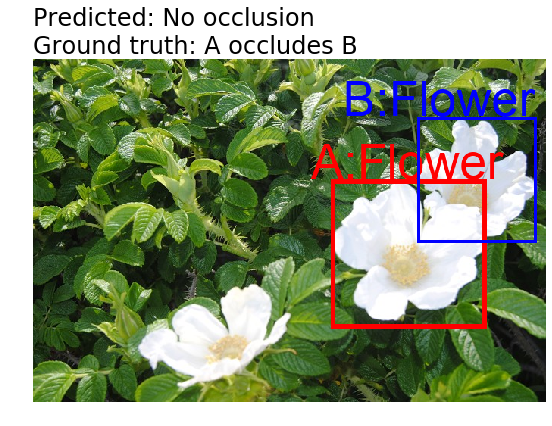}
    \InsertImage{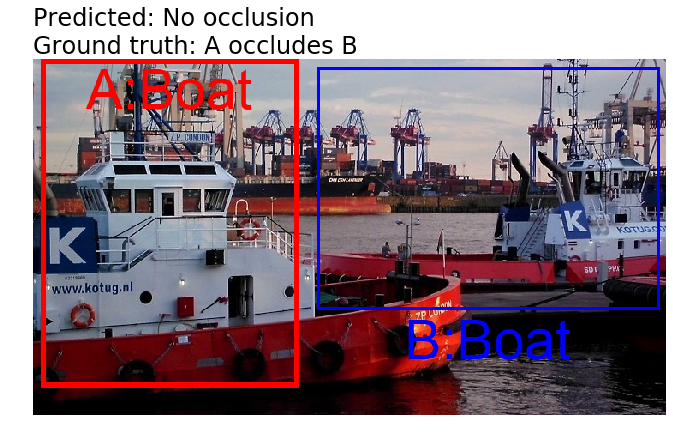}
    \InsertImage{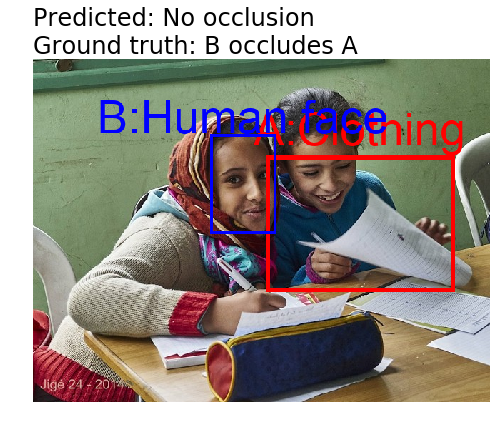}
    \InsertImage{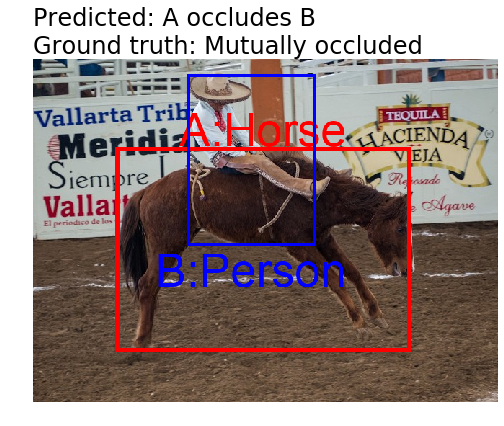}
    \\
    \InsertImage{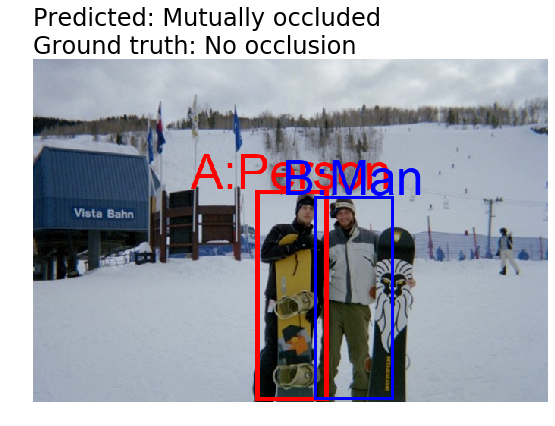}
    \InsertImage{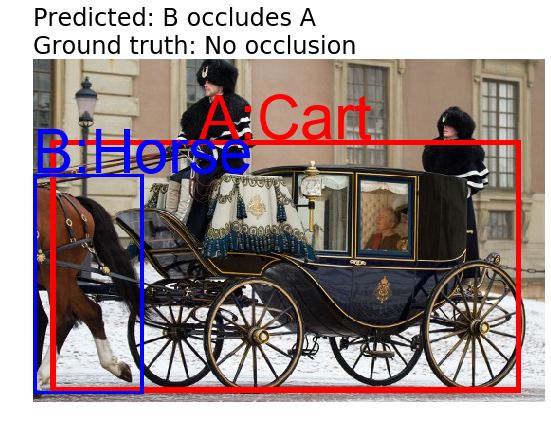}
    \InsertImage{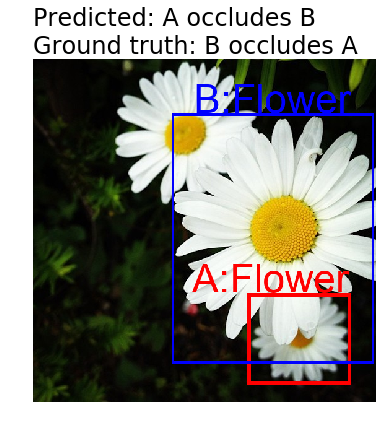}
    \InsertImage{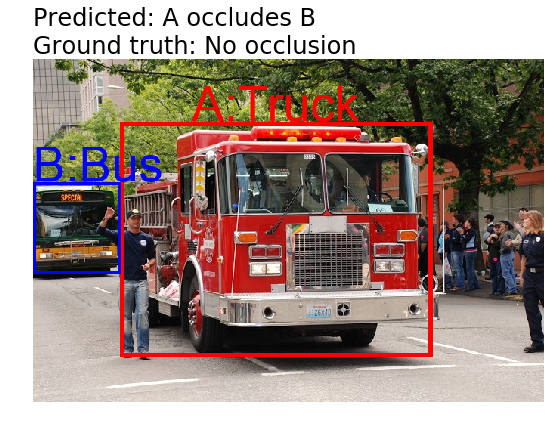}
    \\
    \caption{Failure examples for occlusion prediction.}
    \label{fig:failure_occlusion}
\end{figure*}

\begin{figure*}
    \centering
    \InsertImage[0.48]{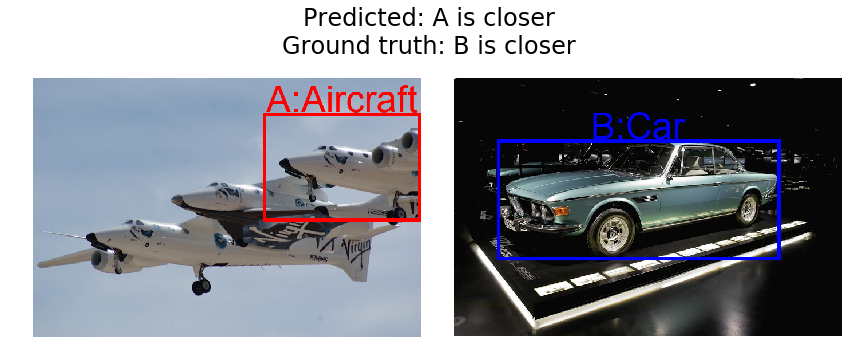}
    \InsertImage[0.48]{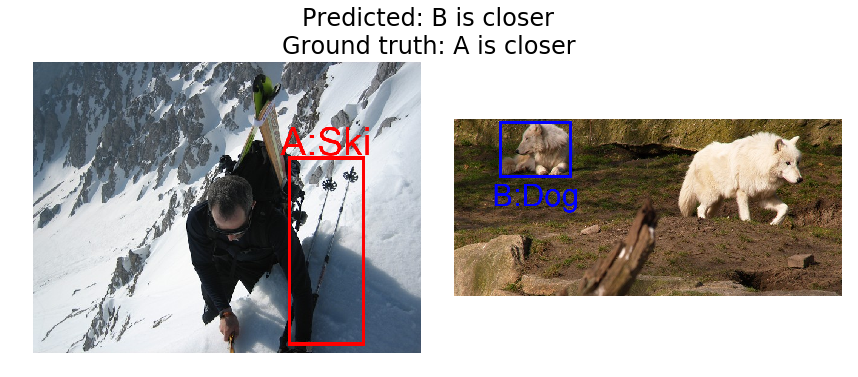}
    \\
    \InsertImage[0.48]{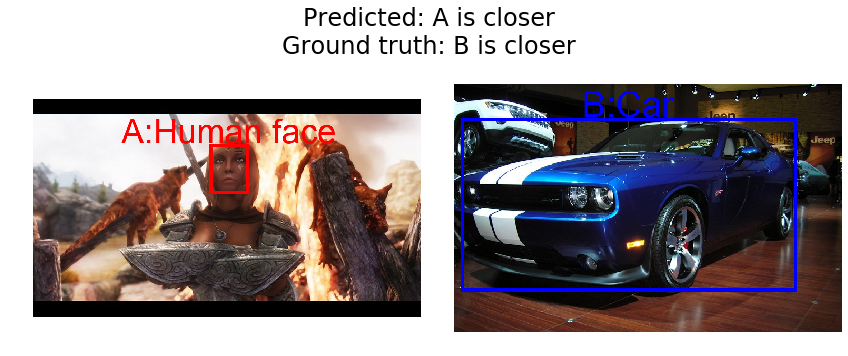}
    \InsertImage[0.48]{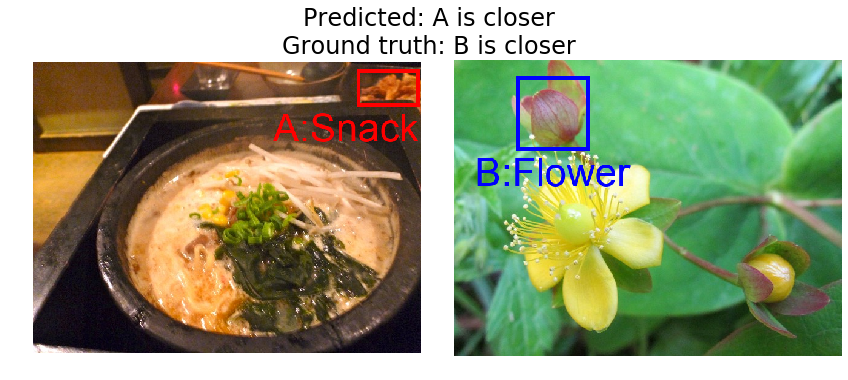}
    \\
    \InsertImage[0.48]{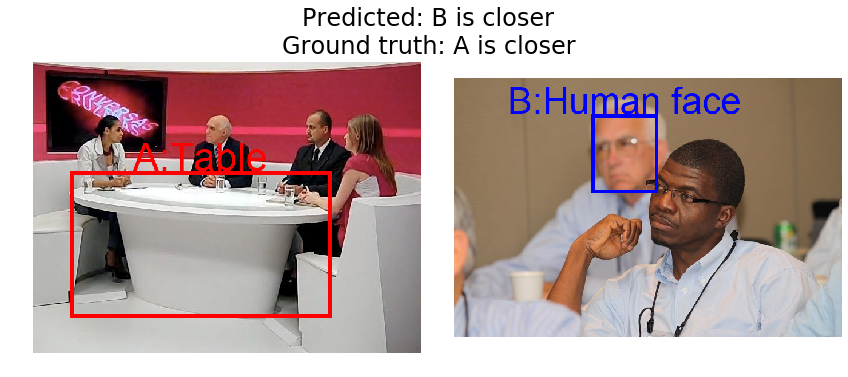}
    \InsertImage[0.48]{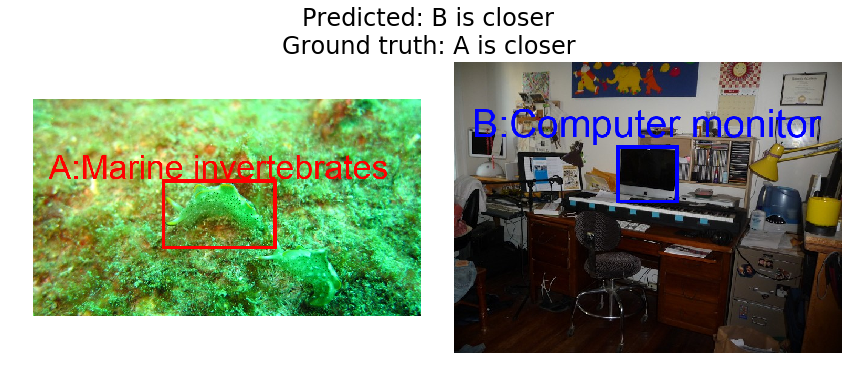}
    \\
    \caption{Failure examples for across-image depth prediction.}
    \label{fig:failure_across}
\end{figure*}

In this section, we present qualitative results of the MLP baseline.
Figure~\ref{fig:qualitative} shows examples for within-image depth relationships with different difficulty scales,
and Figure~\ref{fig:qualitative_occlusion} and Figure~\ref{fig:qualitative_across} are about examples for within-image occlusion and across-image depth, respectively.
We can clearly see the increasing ambiguity in different difficulty levels.
The models manage to differentiate the objects' relative depths for these examples.
We also show failure examples in Figure~\ref{fig:failure_relation}, Figure~\ref{fig:failure_detection}, Figure~\ref{fig:failure_occlusion}, and Figure~\ref{fig:failure_across}.
We can see that the labels may depend on minor differences in objects' depths,
and the models' predictions are reasonable despite that they do not match the groundtruth labels, e.g.,~the last two examples in Figure~\ref{fig:failure_relation}.

{\small
\begin{thebibliography}{10}\itemsep=-1pt

\bibitem{brazil2020kinematic}
Garrick Brazil, Gerard Pons-Moll, Xiaoming Liu, and Bernt Schiele.
\newblock Kinematic 3d object detection in monocular video.
\newblock {\em arXiv preprint arXiv:2007.09548}, 2020.

\bibitem{chao2018learning}
Yu-Wei Chao, Yunfan Liu, Xieyang Liu, Huayi Zeng, and Jia Deng.
\newblock Learning to detect human-object interactions.
\newblock In {\em WACV}, 2018.

\bibitem{chao2015hico}
Yu-Wei Chao, Zhan Wang, Yugeng He, Jiaxuan Wang, and Jia Deng.
\newblock Hico: A benchmark for recognizing human-object interactions in
  images.
\newblock In {\em ICCV}, 2015.

\bibitem{chen2016single}
Weifeng Chen, Zhao Fu, Dawei Yang, and Jia Deng.
\newblock Single-image depth perception in the wild.
\newblock In {\em NeurIPS}, 2016.

\bibitem{Chen_2020_CVPR}
Weifeng Chen, Shengyi Qian, David Fan, Noriyuki Kojima, Max Hamilton, and Jia
  Deng.
\newblock Oasis: A large-scale dataset for single image 3d in the wild.
\newblock In {\em CVPR}, 2020.

\bibitem{chen2016monocular}
Xiaozhi Chen, Kaustav Kundu, Ziyu Zhang, Huimin Ma, Sanja Fidler, and Raquel
  Urtasun.
\newblock Monocular 3d object detection for autonomous driving.
\newblock In {\em CVPR}, 2016.

\bibitem{chen2015multi}
Yi-Ting Chen, Xiaokai Liu, and Ming-Hsuan Yang.
\newblock Multi-instance object segmentation with occlusion handling.
\newblock In {\em CVPR}, 2015.

\bibitem{dai2017detecting}
Bo Dai, Yuqi Zhang, and Dahua Lin.
\newblock Detecting visual relationships with deep relational networks.
\newblock In {\em CVPR}, 2017.

\bibitem{delaitre2011learning}
Vincent Delaitre, Josef Sivic, and Ivan Laptev.
\newblock Learning person-object interactions for action recognition in still
  images.
\newblock In {\em NeurIPS}, 2011.

\bibitem{deng2009imagenet}
Jia Deng, Wei Dong, Richard Socher, Li-Jia Li, Kai Li, and Li Fei-Fei.
\newblock Imagenet: A large-scale hierarchical image database.
\newblock In {\em CVPR}. Ieee, 2009.

\bibitem{deng2017amodal}
Zhuo Deng and Longin Jan~Latecki.
\newblock Amodal detection of 3d objects: Inferring 3d bounding boxes from 2d
  ones in rgb-depth images.
\newblock In {\em CVPR}, 2017.

\bibitem{eigen2015predicting}
David Eigen and Rob Fergus.
\newblock Predicting depth, surface normals and semantic labels with a common
  multi-scale convolutional architecture.
\newblock In {\em ICCV}, 2015.

\bibitem{gao2011segmentation}
Tianshi Gao, Benjamin Packer, and Daphne Koller.
\newblock A segmentation-aware object detection model with occlusion handling.
\newblock In {\em CVPR}, 2011.

\bibitem{geiger2012we}
Andreas Geiger, Philip Lenz, and Raquel Urtasun.
\newblock Are we ready for autonomous driving? the kitti vision benchmark
  suite.
\newblock In {\em CVPR}, 2012.

\bibitem{gkioxari2018detecting}
Georgia Gkioxari, Ross Girshick, Piotr Doll{\'a}r, and Kaiming He.
\newblock Detecting and recognizing human-object interactions.
\newblock In {\em CVPR}, 2018.

\bibitem{goyal2020rel3d}
Ankit Goyal, Kaiyu Yang, Dawei Yang, and Jia Deng.
\newblock Rel3d: A minimally contrastive benchmark for grounding spatial
  relations in 3d.
\newblock {\em Advances in Neural Information Processing Systems 33
  pre-proceedings (NeurIPS)}, 2020.

\bibitem{gupta2009observing}
Abhinav Gupta, Aniruddha Kembhavi, and Larry~S Davis.
\newblock Observing human-object interactions: Using spatial and functional
  compatibility for recognition.
\newblock {\em IEEE Transactions on Pattern Analysis and Machine Intelligence},
  31(10):1775--1789, 2009.

\bibitem{hane2015direction}
Christian Hane, Lubor Ladicky, and Marc Pollefeys.
\newblock Direction matters: Depth estimation with a surface normal classifier.
\newblock In {\em CVPR}, 2015.

\bibitem{hoiem2011recovering}
Derek Hoiem, Alexei~A Efros, and Martial Hebert.
\newblock Recovering occlusion boundaries from an image.
\newblock {\em International Journal of Computer Vision}, 91(3):328--346, 2011.

\bibitem{hsiao2014occlusion}
Edward Hsiao and Martial Hebert.
\newblock Occlusion reasoning for object detectionunder arbitrary viewpoint.
\newblock {\em IEEE Transactions on Pattern Analysis and Machine Intelligence},
  36(9):1803--1815, 2014.

\bibitem{hu2019sail}
Yuan-Ting Hu, Hong-Shuo Chen, Kexin Hui, Jia-Bin Huang, and Alexander~G
  Schwing.
\newblock Sail-vos: Semantic amodal instance level video object segmentation-a
  synthetic dataset and baselines.
\newblock In {\em CVPR}, 2019.

\bibitem{huang2019perspectivenet}
Siyuan Huang, Yixin Chen, Tao Yuan, Siyuan Qi, Yixin Zhu, and Song-Chun Zhu.
\newblock Perspectivenet: 3d object detection from a single rgb image via
  perspective points.
\newblock In {\em NeurIPS}, 2019.

\bibitem{jia2012learning}
Zhaoyin Jia, Andrew Gallagher, Yao-Jen Chang, and Tsuhan Chen.
\newblock A learning-based framework for depth ordering.
\newblock In {\em 2012 IEEE Conference on Computer Vision and Pattern
  Recognition}, pages 294--301. IEEE, 2012.

\bibitem{jiang2020peek}
Ziyu Jiang, Buyu Liu, Samuel Schulter, Zhangyang Wang, and Manmohan Chandraker.
\newblock Peek-a-boo: Occlusion reasoning in indoor scenes with plane
  representations.
\newblock In {\em CVPR}, 2020.

\bibitem{kar2015amodal}
Abhishek Kar, Shubham Tulsiani, Joao Carreira, and Jitendra Malik.
\newblock Amodal completion and size constancy in natural scenes.
\newblock In {\em ICCV}, 2015.

\bibitem{kato2018compositional}
Keizo Kato, Yin Li, and Abhinav Gupta.
\newblock Compositional learning for human object interaction.
\newblock In {\em ECCV}, 2018.

\bibitem{krishna2017visual}
Ranjay Krishna, Yuke Zhu, Oliver Groth, Justin Johnson, Kenji Hata, Joshua
  Kravitz, Stephanie Chen, Yannis Kalantidis, Li-Jia Li, David~A Shamma, et~al.
\newblock Visual genome: Connecting language and vision using crowdsourced
  dense image annotations.
\newblock {\em International Journal of Computer Vision}, 123(1):32--73, 2017.

\bibitem{kuznetsova2018open}
Alina Kuznetsova, Hassan Rom, Neil Alldrin, Jasper Uijlings, Ivan Krasin, Jordi
  Pont-Tuset, Shahab Kamali, Stefan Popov, Matteo Malloci, Tom Duerig, et~al.
\newblock The open images dataset v4: Unified image classification, object
  detection, and visual relationship detection at scale.
\newblock {\em arXiv preprint arXiv:1811.00982}, 2018.

\bibitem{kuznetsova2020open}
Alina Kuznetsova, Hassan Rom, Neil Alldrin, Jasper Uijlings, Ivan Krasin, Jordi
  Pont-Tuset, Shahab Kamali, Stefan Popov, Matteo Malloci, Alexander
  Kolesnikov, et~al.
\newblock The open images dataset v4.
\newblock {\em IJCV}, pages 1--26, 2020.

\bibitem{ladicky2014pulling}
Lubor Ladicky, Jianbo Shi, and Marc Pollefeys.
\newblock Pulling things out of perspective.
\newblock In {\em CVPR}, 2014.

\bibitem{lasinger2019towards}
Katrin Lasinger, Ren{\'e} Ranftl, Konrad Schindler, and Vladlen Koltun.
\newblock Towards robust monocular depth estimation: Mixing datasets for
  zero-shot cross-dataset transfer.
\newblock {\em arXiv preprint arXiv:1907.01341}, 2019.

\bibitem{li2019gs3d}
Buyu Li, Wanli Ouyang, Lu Sheng, Xingyu Zeng, and Xiaogang Wang.
\newblock Gs3d: An efficient 3d object detection framework for autonomous
  driving.
\newblock In {\em CVPR}, 2019.

\bibitem{li2015depth}
Bo Li, Chunhua Shen, Yuchao Dai, Anton Van Den~Hengel, and Mingyi He.
\newblock Depth and surface normal estimation from monocular images using
  regression on deep features and hierarchical crfs.
\newblock In {\em CVPR}, 2015.

\bibitem{li2016amodal}
Ke Li and Jitendra Malik.
\newblock Amodal instance segmentation.
\newblock In {\em ECCV}, 2016.

\bibitem{li2017vip}
Yikang Li, Wanli Ouyang, and Xiaogang Wang.
\newblock Vip-cnn: A visual phrase reasoning convolutional neural network for
  visual relationship detection.
\newblock {\em arXiv preprint arXiv:1702.07191}, 2, 2017.

\bibitem{liang2017deep}
Xiaodan Liang, Lisa Lee, and Eric~P Xing.
\newblock Deep variation-structured reinforcement learning for visual
  relationship and attribute detection.
\newblock In {\em CVPR}, 2017.

\bibitem{lin2014microsoft}
Tsung-Yi Lin, Michael Maire, Serge Belongie, James Hays, Pietro Perona, Deva
  Ramanan, Piotr Doll{\'a}r, and C~Lawrence Zitnick.
\newblock Microsoft coco: Common objects in context.
\newblock In {\em ECCV}. Springer, 2014.

\bibitem{liu2010single}
Beyang Liu, Stephen Gould, and Daphne Koller.
\newblock Single image depth estimation from predicted semantic labels.
\newblock In {\em CVPR}, 2010.

\bibitem{liu2015deep}
Fayao Liu, Chunhua Shen, and Guosheng Lin.
\newblock Deep convolutional neural fields for depth estimation from a single
  image.
\newblock In {\em CVPR}, 2015.

\bibitem{liu2019deep}
Lijie Liu, Jiwen Lu, Chunjing Xu, Qi Tian, and Jie Zhou.
\newblock Deep fitting degree scoring network for monocular 3d object
  detection.
\newblock In {\em CVPR}, 2019.

\bibitem{lu2016visual}
Cewu Lu, Ranjay Krishna, Michael Bernstein, and Li Fei-Fei.
\newblock Visual relationship detection with language priors.
\newblock In {\em ECCV}, 2016.

\bibitem{lu2019occlusion}
Rui Lu, Feng Xue, Menghan Zhou, Anlong Ming, and Yu Zhou.
\newblock Occlusion-shared and feature-separated network for occlusion
  relationship reasoning.
\newblock In {\em Proceedings of the IEEE/CVF International Conference on
  Computer Vision}, pages 10343--10352, 2019.

\bibitem{ma2018attend}
Chih-Yao Ma, Asim Kadav, Iain Melvin, Zsolt Kira, Ghassan AlRegib, and Hans
  Peter~Graf.
\newblock Attend and interact: Higher-order object interactions for video
  understanding.
\newblock In {\em CVPR}, 2018.

\bibitem{ma2019accurate}
Xinzhu Ma, Zhihui Wang, Haojie Li, Pengbo Zhang, Wanli Ouyang, and Xin Fan.
\newblock Accurate monocular 3d object detection via color-embedded 3d
  reconstruction for autonomous driving.
\newblock In {\em ICCV}, 2019.

\bibitem{mousavian20173d}
Arsalan Mousavian, Dragomir Anguelov, John Flynn, and Jana Kosecka.
\newblock 3d bounding box estimation using deep learning and geometry.
\newblock In {\em CVPR}, 2017.

\bibitem{peyre2017weakly}
Julia Peyre, Josef Sivic, Ivan Laptev, and Cordelia Schmid.
\newblock Weakly-supervised learning of visual relations.
\newblock In {\em ICCV}, 2017.

\bibitem{plummer2017phrase}
Bryan~A Plummer, Arun Mallya, Christopher~M Cervantes, Julia Hockenmaier, and
  Svetlana Lazebnik.
\newblock Phrase localization and visual relationship detection with
  comprehensive image-language cues.
\newblock In {\em ICCV}, 2017.

\bibitem{qi2019amodal}
Lu Qi, Li Jiang, Shu Liu, Xiaoyong Shen, and Jiaya Jia.
\newblock Amodal instance segmentation with kins dataset.
\newblock In {\em CVPR}, 2019.

\bibitem{qi2018learning}
Siyuan Qi, Wenguan Wang, Baoxiong Jia, Jianbing Shen, and Song-Chun Zhu.
\newblock Learning human-object interactions by graph parsing neural networks.
\newblock In {\em ECCV}, 2018.

\bibitem{ren2015faster}
Shaoqing Ren, Kaiming He, Ross Girshick, and Jian Sun.
\newblock Faster r-cnn: Towards real-time object detection with region proposal
  networks.
\newblock In {\em NeurIPS}, 2015.

\bibitem{humanvision}
BJ~Rogers RJ~Watt.
\newblock Human vision and cognitive science.
\newblock {\em Research Directions in Cognitive Science: A European
  Perspective}, 1989.

\bibitem{russell2008labelme}
Bryan~C Russell, Antonio Torralba, Kevin~P Murphy, and William~T Freeman.
\newblock Labelme: a database and web-based tool for image annotation.
\newblock {\em International journal of computer vision}, 77(1-3):157--173,
  2008.

\bibitem{sadeghi2011recognition}
Mohammad~Amin Sadeghi and Ali Farhadi.
\newblock Recognition using visual phrases.
\newblock In {\em CVPR}, 2011.

\bibitem{saxena2006learning}
Ashutosh Saxena, Sung~H Chung, and Andrew~Y Ng.
\newblock Learning depth from single monocular images.
\newblock In {\em NeurIPS}, 2006.

\bibitem{saxena2008make3d}
Ashutosh Saxena, Min Sun, and Andrew~Y Ng.
\newblock Make3d: Learning 3d scene structure from a single still image.
\newblock {\em IEEE Transactions on Pattern Analysis and Machine Intelligence},
  31(5):824--840, 2008.

\bibitem{shelhamer2015scene}
Evan Shelhamer, Jonathan~T Barron, and Trevor Darrell.
\newblock Scene intrinsics and depth from a single image.
\newblock In {\em ICCV Workshops}, 2015.

\bibitem{silberman2012indoor}
Nathan Silberman, Derek Hoiem, Pushmeet Kohli, and Rob Fergus.
\newblock Indoor segmentation and support inference from rgbd images.
\newblock In {\em ECCV}, 2012.

\bibitem{song2015sun}
Shuran Song, Samuel~P Lichtenberg, and Jianxiong Xiao.
\newblock Sun rgb-d: A rgb-d scene understanding benchmark suite.
\newblock In {\em CVPR}, 2015.

\bibitem{sundberg2011occlusion}
Patrik Sundberg, Thomas Brox, Michael Maire, Pablo Arbel{\'a}ez, and Jitendra
  Malik.
\newblock Occlusion boundary detection and figure/ground assignment from
  optical flow.
\newblock In {\em CVPR 2011}, pages 2233--2240. IEEE, 2011.

\bibitem{szegedy2016inception}
Christian Szegedy, Sergey Ioffe, Vincent Vanhoucke, and Alex Alemi.
\newblock Inception-v4, inception-resnet and the impact of residual connections
  on learning.
\newblock {\em arXiv preprint arXiv:1602.07261}, 2016.

\bibitem{tighe2014scene}
Joseph Tighe, Marc Niethammer, and Svetlana Lazebnik.
\newblock Scene parsing with object instances and occlusion ordering.
\newblock In {\em CVPR}, 2014.

\bibitem{weiss2001model}
Isaac Weiss and Manjit Ray.
\newblock Model-based recognition of 3d objects from single images.
\newblock {\em IEEE Transactions on Pattern Analysis and Machine Intelligence},
  23(2):116--128, 2001.

\bibitem{woo2018linknet}
Sanghyun Woo, Dahun Kim, Donghyeon Cho, and In~So Kweon.
\newblock Linknet: Relational embedding for scene graph.
\newblock In {\em NeurIPS}, 2018.

\bibitem{xian2018monocular}
Ke Xian, Chunhua Shen, Zhiguo Cao, Hao Lu, Yang Xiao, Ruibo Li, and Zhenbo Luo.
\newblock Monocular relative depth perception with web stereo data supervision.
\newblock In {\em CVPR}, 2018.

\bibitem{xian2020structure}
Ke Xian, Jianming Zhang, Oliver Wang, Long Mai, Zhe Lin, and Zhiguo Cao.
\newblock Structure-guided ranking loss for single image depth prediction.
\newblock In {\em CVPR}, 2020.

\bibitem{xiang2014beyond}
Yu Xiang, Roozbeh Mottaghi, and Silvio Savarese.
\newblock Beyond pascal: A benchmark for 3d object detection in the wild.
\newblock In {\em WACV}, 2014.

\bibitem{yang2018graph}
Jianwei Yang, Jiasen Lu, Stefan Lee, Dhruv Batra, and Devi Parikh.
\newblock Graph r-cnn for scene graph generation.
\newblock In {\em ECCV}, 2018.

\bibitem{yang2019spatialsense}
Kaiyu Yang, Olga Russakovsky, and Jia Deng.
\newblock Spatialsense: An adversarially crowdsourced benchmark for spatial
  relation recognition.
\newblock In {\em ICCV}, 2019.

\bibitem{yao2010grouplet}
Bangpeng Yao and Li Fei-Fei.
\newblock Grouplet: A structured image representation for recognizing human and
  object interactions.
\newblock In {\em CVPR}, 2010.

\bibitem{yao2018exploring}
Ting Yao, Yingwei Pan, Yehao Li, and Tao Mei.
\newblock Exploring visual relationship for image captioning.
\newblock In {\em ECCV}, 2018.

\bibitem{yin2018zoom}
Guojun Yin, Lu Sheng, Bin Liu, Nenghai Yu, Xiaogang Wang, Jing Shao, and Chen
  Change~Loy.
\newblock Zoom-net: Mining deep feature interactions for visual relationship
  recognition.
\newblock In {\em ECCV}, 2018.

\bibitem{yu2017visual}
Ruichi Yu, Ang Li, Vlad~I Morariu, and Larry~S Davis.
\newblock Visual relationship detection with internal and external linguistic
  knowledge distillation.
\newblock In {\em ICCV}, 2017.

\bibitem{zhan2020self}
Xiaohang Zhan, Xingang Pan, Bo Dai, Ziwei Liu, Dahua Lin, and Chen~Change Loy.
\newblock Self-supervised scene de-occlusion.
\newblock In {\em CVPR}, 2020.

\bibitem{zhang2017visual}
Hanwang Zhang, Zawlin Kyaw, Shih-Fu Chang, and Tat-Seng Chua.
\newblock Visual translation embedding network for visual relation detection.
\newblock In {\em CVPR}, 2017.

\bibitem{zhang2017ppr}
Hanwang Zhang, Zawlin Kyaw, Jinyang Yu, and Shih-Fu Chang.
\newblock Ppr-fcn: Weakly supervised visual relation detection via parallel
  pairwise r-fcn.
\newblock In {\em ICCV}, 2017.

\bibitem{zhang2017relationship}
Ji Zhang, Mohamed Elhoseiny, Scott Cohen, Walter Chang, and Ahmed Elgammal.
\newblock Relationship proposal networks.
\newblock In {\em CVPR}, 2017.

\bibitem{zhang2019large}
Ji Zhang, Yannis Kalantidis, Marcus Rohrbach, Manohar Paluri, Ahmed Elgammal,
  and Mohamed Elhoseiny.
\newblock Large-scale visual relationship understanding.
\newblock In {\em AAAI}, 2019.

\bibitem{zhang2018occlusion}
Shifeng Zhang, Longyin Wen, Xiao Bian, Zhen Lei, and Stan~Z Li.
\newblock Occlusion-aware r-cnn: detecting pedestrians in a crowd.
\newblock In {\em ECCV}, 2018.

\bibitem{zhang2015monocular}
Ziyu Zhang, Alexander~G Schwing, Sanja Fidler, and Raquel Urtasun.
\newblock Monocular object instance segmentation and depth ordering with cnns.
\newblock In {\em Proceedings of the IEEE International Conference on Computer
  Vision}, pages 2614--2622, 2015.

\bibitem{zhu2017semantic}
Yan Zhu, Yuandong Tian, Dimitris Metaxas, and Piotr Doll{\'a}r.
\newblock Semantic amodal segmentation.
\newblock In {\em Proceedings of the IEEE Conference on Computer Vision and
  Pattern Recognition}, pages 1464--1472, 2017.

\bibitem{zhuang2017towards}
Bohan Zhuang, Lingqiao Liu, Chunhua Shen, and Ian Reid.
\newblock Towards context-aware interaction recognition for visual relationship
  detection.
\newblock In {\em ICCV}, 2017.

\bibitem{zhuang2017care}
Bohan Zhuang, Qi Wu, Chunhua Shen, Ian Reid, and Anton van~den Hengel.
\newblock Care about you: towards large-scale human-centric visual relationship
  detection.
\newblock {\em arXiv preprint arXiv:1705.09892}, 2017.

\end{thebibliography}

}
\end{document}